\documentclass[letterpaper]{article} 
\usepackage{aaai24}  
\usepackage{times}  
\usepackage{helvet}  
\usepackage{courier}  
\usepackage[hyphens]{url}  
\usepackage{graphicx} 
\usepackage{natbib}  
\usepackage{caption} 
\usepackage{algorithm}
\usepackage{algorithmic}

\usepackage{times}
\usepackage{epsfig}
\usepackage{graphicx}
\usepackage{amsmath}
\usepackage{amssymb}
\usepackage{booktabs}
\usepackage{multirow}
\usepackage{multicol}
\usepackage[table,xcdraw]{xcolor}
\usepackage{pifont}
\usepackage{isotope}
\newcommand\thefont{\expandafter\string\the\font}

\newcolumntype{L}[1]{>{\raggedright\let\newline\\\arraybackslash\hspace{0pt}}m{#1}}
\newcolumntype{C}[1]{>{\centering\let\newline\\\arraybackslash\hspace{0pt}}m{#1}}
\newcolumntype{R}[1]{>{\raggedleft\let\newline\\\arraybackslash\hspace{0pt}}m{#1}}

\usepackage{subcaption}
\usepackage{newfloat}
\usepackage{listings}
\DeclareCaptionStyle{ruled}{labelfont=normalfont,labelsep=colon,strut=off} 
\floatstyle{ruled}
\newfloat{listing}{tb}{lst}{}
\floatname{listing}{Listing}
\usepackage{amssymb}
\usepackage{amsmath}
\usepackage{booktabs}
\usepackage{multirow}
\usepackage{pifont}
\newcommand{\cmark}{\ding{51}}

\usepackage[capitalize]{cleveref}
\crefname{section}{Sec.}{Secs.}
\Crefname{section}{Section}{Sections}
\Crefname{table}{Table}{Tables}
\crefname{table}{Tab.}{Tabs.}
\Crefname{equation}{Equation}{Equations}
\crefname{equation}{Eq.}{Eqs.}

\title{VVS: Video-to-Video Retrieval with Irrelevant Frame Suppression}
\author {
    Won Jo\textsuperscript{\rm 1},
    Geuntaek Lim\textsuperscript{\rm 1},
    Gwangjin Lee\textsuperscript{\rm 1},
    Hyunwoo Kim\textsuperscript{\rm 1},
    Byungsoo Ko\textsuperscript{\rm 2},
    and Yukyung Choi\textsuperscript{\rm 1}
}
\affiliations {
    \textsuperscript{\rm 1}Sejong University\\
    \textsuperscript{\rm 2}NAVER Vision\\
    \{jwon, gtlim, gjlee, hwkim\}@rcv.sejong.ac.kr,  kobiso62@gmail.com, ykchoi@rcv.sejong.ac.kr
}

\begin{document}

\maketitle

\begin{abstract}
    In content-based video retrieval (CBVR), dealing with large-scale collections, efficiency is as important as accuracy; thus, several video-level feature-based studies have actively been conducted. Nevertheless, owing to the severe difficulty of embedding a lengthy and untrimmed video into a single feature, these studies have been insufficient for accurate retrieval compared to frame-level feature-based studies. In this paper, we show that appropriate suppression of irrelevant frames can provide insight into the current obstacles of the video-level approaches. Furthermore, we propose a Video-to-Video Suppression network (VVS) as a solution. VVS is an end-to-end framework that consists of an easy distractor elimination stage to identify which frames to remove and a suppression weight generation stage to determine the extent to suppress the remaining frames. This structure is intended to effectively describe an untrimmed video with varying content and meaningless information. Its efficacy is proved via extensive experiments, and we show that our approach is not only state-of-the-art in video-level approaches but also has a fast inference time despite possessing retrieval capabilities close to those of frame-level approaches. Code is available at https://github.com/sejong-rcv/VVS
\end{abstract}

\section{Introduction}
    Information retrieval is defined as finding the most relevant information in a large collection. It has evolved from finding text within a document~\cite{griffiths1986using, strzalkowski1995natural, bellot1999clustering, liu2004cluster} to finding images within an image set~\cite{arandjelovic2016netvlad, tolias2016particular, jun2019combination, ko2019benchmark, ko2020embedding, gu2020symmetrical}. In recent years, with the fast growing trend of the video streaming market, several studies~\cite{kordopatis2017near, kordopatis2019visil, shao2021temporal, jo2022exploring, ng2022vrag} have actively been conducted in content-based video retrieval (CBVR) to find desired videos from a set of videos.

    The core of CBVR technology is to measure similarities between videos of different lengths, including untrimmed videos. This is divided into two streams according to the basic unit for measuring the similarity between two videos: a frame-level feature-based approach and a video-level feature-based approach. The former aggregates similarities between frame-level features in two videos to calculate a video-to-video similarity. Conversely, the latter describes each video as a single feature and computes a video-to-video similarity based on it. These two streams are in a trade-off relationship because the key foundation for determining similarity differs. The frame-level approach compares each frame directly; it is less dependent on factors such as video duration and whether or not it is trimmed. As a result, relatively accurate searches are possible, but processing speed and memory are expensive due to the necessity of numerous similarity computations and a considerable amount of feature storage space. In comparison, the video-level approach requires only one similarity calculation between a single pair of features, which is more efficient in terms of processing speed and memory. However, it is difficult to compress many frames of a video into a single feature, making approaches of this type generally inaccurate and sensitive to factors such as duration and trimness.

    \begin{figure}[t!]
        \centering
        \includegraphics[width=0.925\linewidth]{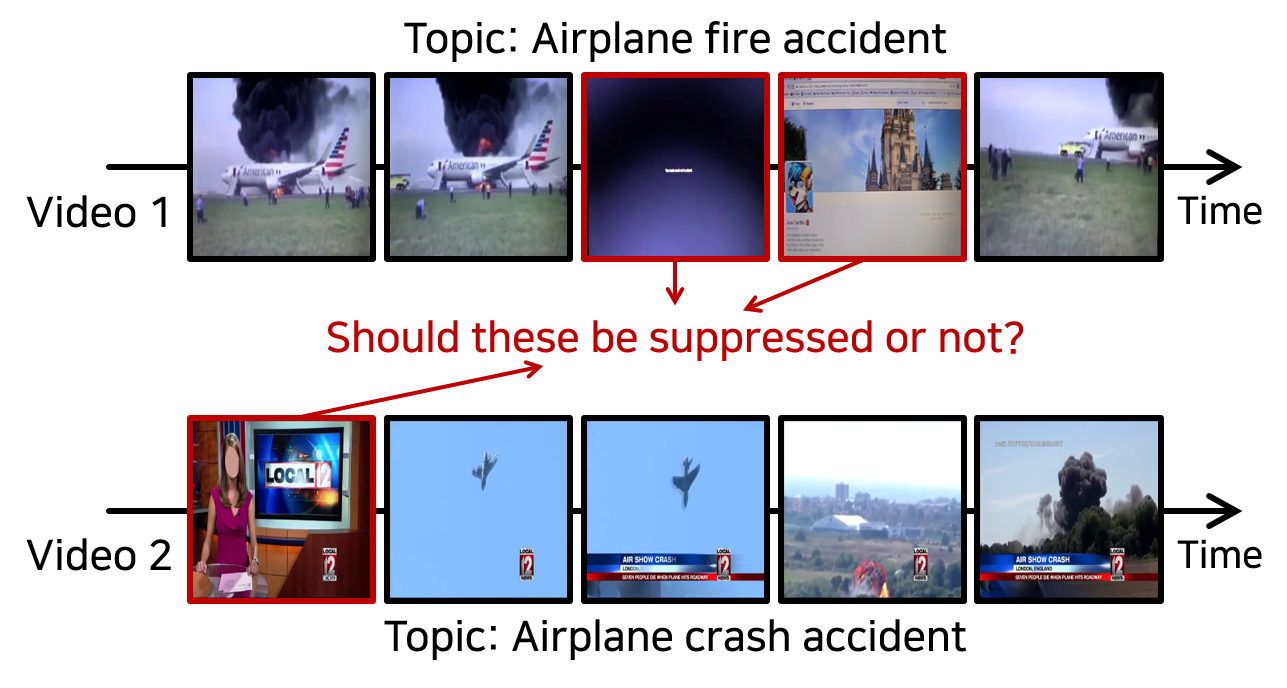} 
        \caption{\textbf{Q: Should the red boxes be suppressed?} The red boxes in both videos should be excluded because they are unrelated to the topic in the video, although they are included for a specific purpose or reason. In this work, we demonstrate that the suppression of these red boxes enhances the distinctiveness of features when describing the entire video at once. } \label{fig:fig1}
    \end{figure}
   
    Ideally, if a video-level approach could be as distinct as a frame-level approach, it may be the best option in real-world scenarios. However, there are some problems that must be considered. First is that distractors in a video interfere with the description of video-level features. Distractors in this context refer to frames with visual content that is unrelated to the main topic. Indeed, as shown in the two video examples in~\Cref{fig:fig1}, it is obvious that the red box frames corresponding to the distractors are not helpful for recognizing the topic of each video. We also present an experiment that demonstrates quantitative performance improvements when the distractors are manually eliminated from the previous video-level feature-based schemes in the supplementary material. On the basis of these observations, this study proves that the description of video-level features with optimal suppression of distractors can be an ideal scenario for accurate and fast retrieval.

    \begin{figure}[t!]
        \centering
        \includegraphics[width=0.925\linewidth]{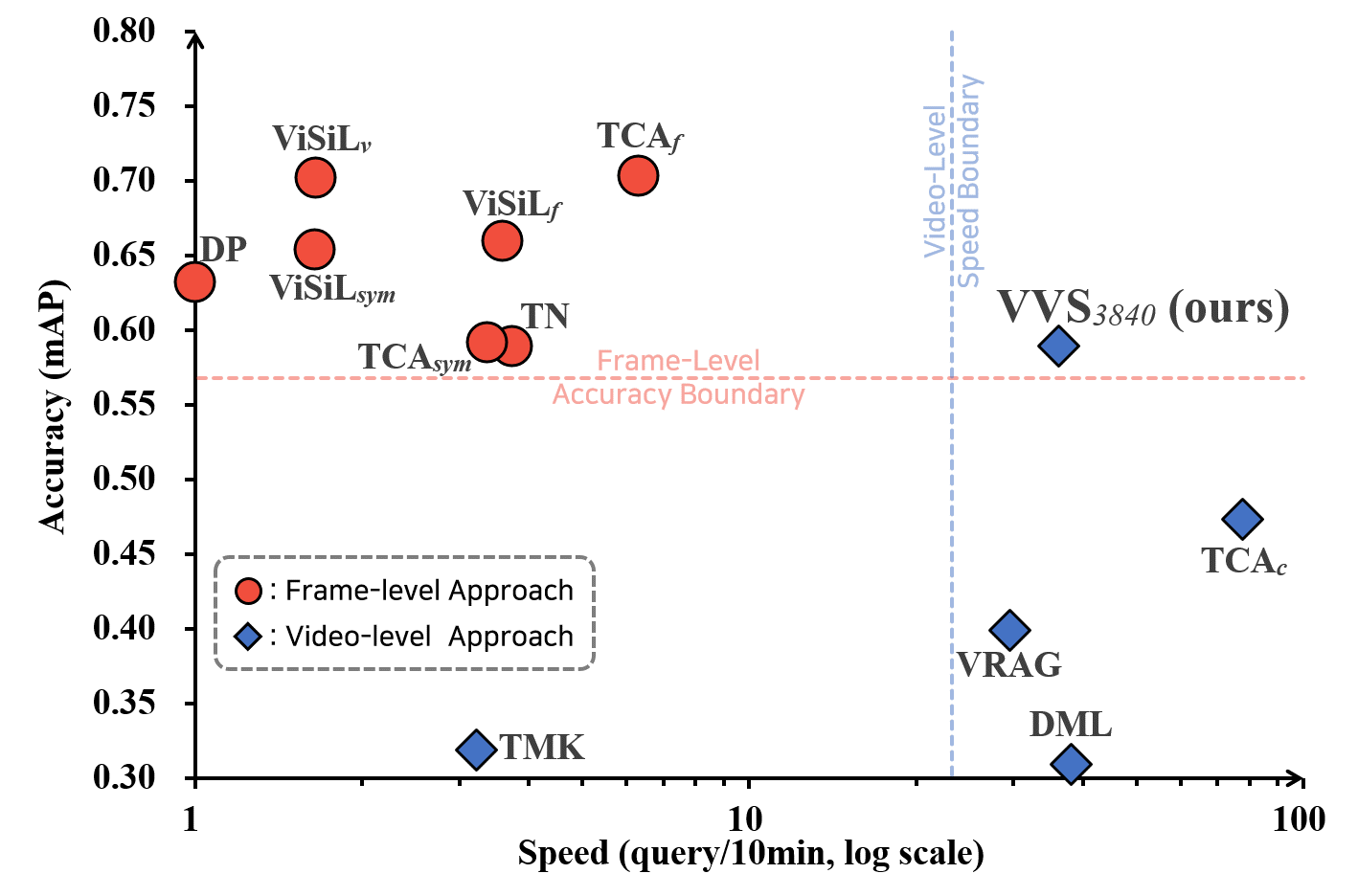}
        \caption{\textbf{Speed-Accuracy Comparison on FIVR-200K.} This is a comparison between the proposed approach and existing state-of-the-art approaches in terms of speed and accuracy on the FIVR-200K. Speed is represented by the average number of queries processed in 10 minutes, and accuracy is represented by the mAP in ISVR, the most difficult task.} \label{fig:spd_acc}
    \end{figure}

    The objective of this work is to understand the significance of frames to determine how much they should be suppressed to produce a distinct video-level feature. To this end, we propose a Video-to-Video Suppression network~(VVS). The VVS is an end-to-end framework consisting of two stages: an easy distractor elimination stage for removing frames that can be clearly recognized as distractors, and a suppression weight generation stage for determining how much to suppress the remaining frames via temporal saliency information and relevance of the topic. Our solution is the first explicitly designed framework that employs various signals for relevance, as opposed to earlier approaches~\cite{kordopatis2017near, shao2021temporal, ng2022vrag} where the model was implicitly intended to generate weights. As shown in~\Cref{fig:spd_acc}, VVS achieves state-of-the-art performance among video-level approaches, with search accuracy comparable to frame-level state-of-the-art performance while retaining competitive inference speed. In addition, extensive experiments included in the later section demonstrate the effectiveness of the proposed framework and the validity of the designed structure.
    
    In summary, our main contribution is as follows: 1)~we demonstrate that video-level features can be both accurate and fast with proper suppression of irrelevant frames, 2) we propose VVS, an end-to-end framework for embedding an untrimmed video as a video-level feature while suppressing frames via various signals, and 3) we show extensive experiments that demonstrate the effectiveness of our design, which acquires state-of-the-art performance.

\section{Related Work} 
        
    \subsection{Frame-level Feature-based Approaches}
        There have been several recent studies in frame-level feature-based approaches. Dynamic Programming (DP) \cite{chou2015pattern} detects a near-duplicate region by extracting the diagonal pattern from a frame-level similarity map. Temporal Network (TN)~\cite{tan2009scalable} distinguishes the longest route in a graph created by keypoint frame matching to discover visually similar frames between two videos, and Circulant Temporal Encoding (CTE)~\cite{douze2016circulant} compares frame-level features using a Fourier transform. This allows frame information to be encoded in the frequency domain. The Video Similarity Learning (ViSiL)~\cite{kordopatis2019visil} approach leverages metric learning by basing its operations on a frame-by-frame similarity map, while Temporal Nested Invariance Pooling~\cite{jo2022exploring} uses a local context-invariant property to design temporally robust pooling based on the standard~\cite{cfp_cdva}. These approaches have higher accuracy than existing video-level approaches, but they are significantly slower in terms of search speed.
        
    \subsection{Video-level Feature-based Approaches}
        Various video-level approaches have also been explored in recent studies. Hashing Code (HC)~\cite{song2013effective} collects and hashes a large number of local and global features to handle accuracy and scalability issues. Deep Metric Learning (DML)~\cite{kordopatis2017near} utilizes frame-level features from a layer codebook generated for intermediate Maximum Activation of Convolution (iMAC)~\cite{kordopatis2017nearimac} features and fuses them to represent a video-level feature. Temporal Matching Kernel (TMK)~\cite{poullot2015temporal} generates a fixed length sequence for each video, regardless of the total number of frames in the video, using periodic kernels that take into account frame descriptors and timestamps. Furthermore, Learning to Align and Match Videos (LAMV)~\cite{baraldi2018lamv} designs a learnable feature transform coefficient based on TMK. Temporal Context Aggregation (TCA)~\cite{shao2021temporal} learns frame-level features into video-level features through self-attention and a queue-based training mechanism, while Distill-and-Select (DNS)~\cite{kordopatis2022dns} distills the knowledge of the teacher network, which is optimized from the labeled data, into a fine or coarse-grained student network to take further advantage of learning from the unlabeled data. This approach also maintains efficiency between the two types of students via a selector network. Video Region Attention Graph (VRAG)~\cite{ng2022vrag} learns an embedding for a video by capturing the relationship of region units in frames via graph attention~\cite{velivckovic2017graph} layers.
        
        In general, these approaches can respond to a given query more quickly than frame-level approaches, even if the response is relatively inaccurate. However, our solution can respond as precisely as frame-level approaches while maintaining sufficient speed as a video-level approach. In addition, whereas DML, TCA, and VRAG (the most similar approaches to ours) ask FC layers, self-attention layers, and graph attention layers, respectively, to implicitly fuse frame-level features into a video-level feature (that is, only contrastive loss of fused features is used as the objective function), our approach is the first to generate video-level features via explicit signals, such as low-level characteristics, temporal saliency, and rough topic.
        
\section{Approach}
    
    \subsection{Problem Formulation}
\begin{figure}[t!]
        \centering
        \includegraphics[width=0.93\linewidth]{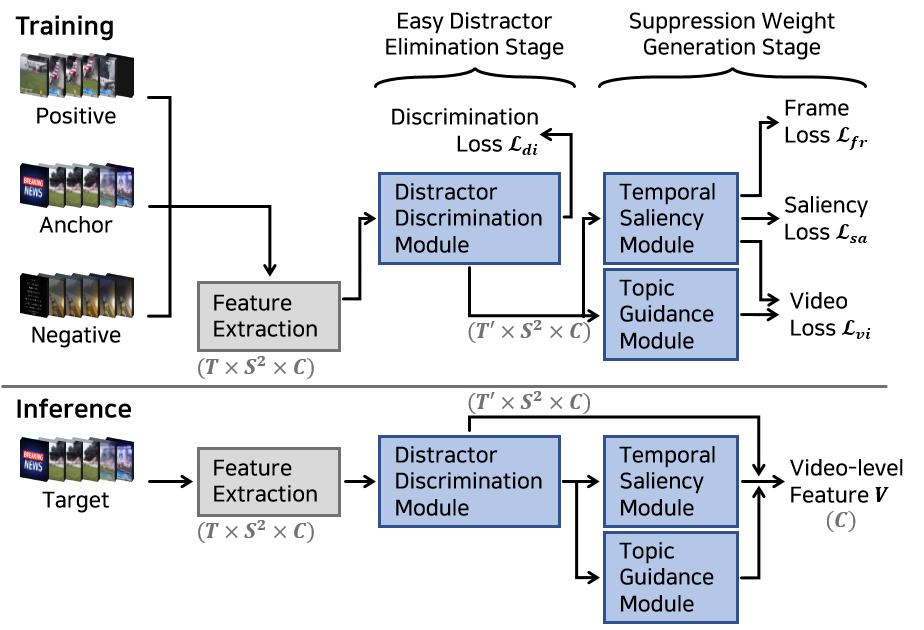} 
        \caption{\textbf{Pipeline Overview of VVS.} The gray italic letters represent the size of the feature in each process.} \label{fig:overview}
    \end{figure}

        Given a video with a duration of $T$, our goal is to embed it as a video-level feature~$V$ while suppressing frames corresponding to distractors. To determine which frame and to what extent it should be suppressed, video frames are first embedded in the frame-level features~$X$~$=$~$\{ x^{(t)} \}_{t=1}^{T}$ instead of being embedded directly in the video-level feature~$V$. Next, $T'$ frames are chosen by removing easy distractors that are readily identifiable as distractors due to a lack of information via the easy distractor elimination stage. In the subsequent suppression weight generation stage, weights $W$~$=$~$\{ w^{(t)} \}_{t=1}^{T'}$ indicating the necessary degree of the remaining frames are calculated. Consequently, these weights are used to aggregate frame-level features into a video-level feature~$V$~$=$~$\Psi(\{w^{(t)}\otimes x^{(t)}\}_{t=1}^{T'})$, where $\Psi$ represents the Spatio-Temporal Global Average Pooling (ST-GAP) and $\otimes$ represents the Hadamard product. \Cref{fig:overview} illustrates an overview of the VVS pipeline.

    \subsection{Feature Extraction}

        $\textup{L}_{N}\textup{-iMAC}$ as a frame-level feature is first extracted for fair comparisons with many other works~\cite{kordopatis2017near, kordopatis2019visil, shao2021temporal, ng2022vrag}. Specifically, each frame is fed to the backbone network $\Phi$ as input, and Regional Maximum Activation of Convolution (R-MAC)~\cite{tolias2016particular} is applied to its intermediate feature maps~$\cal{M}^{\mathit{(k)}}$~$\in$~$\mathbb{R}^{\mathit{S^\mathrm{2} \times C^{\mathit{(k)}} (k=1,\cdots,K)}}$. Specifically, after obtaining feature maps from a total of $K$ layers in $\Phi$, $N$ types of region kernels are used, depending on the granularity level, for applying R-MAC. As a result, each of the $K$ intermediate feature maps $\cal{M}^{\mathit{(k)}}$ have their own channel~${C^{(k)}}$ but the same spatial resolution~$S^2$. After these~$\cal{M}^{\mathit{(k)}}$ are concatenated on the channel axis, they are generated as a frame-level feature $x$~$\in$~$\mathbb{R}^{S^2 \times C \,(C=\sum_{k=1}^{K}C^{(k)})}$. After applying Principal Component Analysis~(PCA) whitening~\cite{jegou2012negative} to each of the features~$x$, the $\textup{L}_{N}\textup{-iMAC}$ feature~$X$~$\in $~$\mathbb{R}^{T \times S^2 \times C}$ is obtained. Although the dimension of the channel axis could be reduced to different sizes for comparison with other approaches when applying PCA whitening, for convenience, the dimension of the frame-level feature is called $C$.

    \subsection{Easy Distractor Elimination Stage}
\begin{figure}[t!]     
            \centering
            \includegraphics[width=0.93\linewidth]{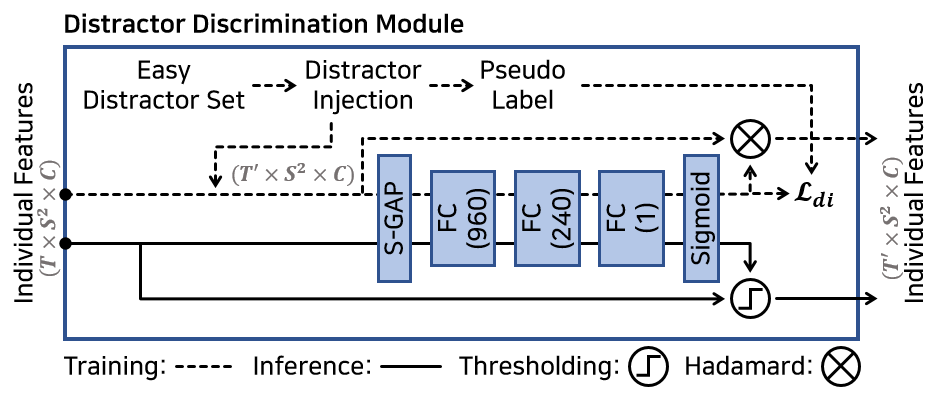} 
            \caption{\textbf{Pipeline of DDM.} The gray italic letters indicate the size of the feature in each process. The number in parentheses in the layer blocks indicates the output dimension.} \label{fig:ddm}
        \end{figure}

        In this section, we introduce the Distractor Discrimination Module (DDM), which eliminates frames that are clearly recognizable as distractors due to a lack of visual information. An easy distractor is a frame with little variation in pixel intensity and few low-level characteristics~(edges, corners, etc.) in an image, such as the third frame of the first video in \Cref{fig:fig1}. In the training phase of DDM, frame-level features corresponding to the easy distractor are injected into an input with a length of $T$, and the model is optimized to distinguish them. In the inference phase of DDM, frames predicted as easy distractors are removed from the input. This process results in the output length being longer than the input length $T$ in the training phase but shorter in the inference phase. For convenience, the output length of DDM is always called $T'$. The overall flow is depicted in~\Cref{fig:ddm}.
        
        \paragraph{Distractor Discrimination Module \\}
            To enable this module to learn to recognize an easy distractor, pseudo-labels are created using the magnitude of the frame-level features. This is because frames with few low-level characteristics have fewer elements to be activated from the backbone network of $\textup{L}_{N}\textup{-iMAC}$, which consists of several activation layers, resulting in a smaller magnitude of their intermediate feature map. 
            
            Specifically, before the training phase, a set of easy distractors with a magnitude lower than or equal to a magnitude threshold~$\lambda_{mag}$ is constructed from all frame-level features of the videos in the training dataset. Examples of easy distractors included in this set can be found in the supplementary material. During the training phase, features of easy distractors are picked from the set and randomly placed between the features~$X$. In this case, only about $20$–$50\%$ of $T$ are injected, resulting in features of length $T'$. Simultaneously, the points where the distractors are injected are set at $0$ and the opposite position at~$1$, resulting in a pseudo-label~$Y_{di}$~$=$~$\{ y^{(t)}_{di}  \}^{T'}_{t=1}$. The injected features are projected through multiple layers to calculate a confidence $W_{di}$~$=$~$\{ w^{(t)}_{di} \}^{T'}_{t=1}$. Because only the components within each frame determine the criterion for identifying easy distractors, the multiple layers consist of only the Spatial Global Average Pooling (S-GAP) and FC layers to handle each frame independently without interaction between frames. As a result, this module is optimized by discrimination loss $\mathcal{L}_{di}$, computed as the binary cross entropy loss between the confidence $W_{di}$ and the pseudo-label~$Y_{di}$. 
            
            The objective of DDM is to convey features to the subsequent stage, erasing features of frames that are deemed to be easy distractors through thresholding for confidence. In this case, since the threshold operation is not differentiable during the training phase, the output is derived from the Hadamard product of the confidence~$W_{di}$ and the input features $X$, and during the inference phase, from a thresholding operation based on a distractor threshold $\lambda_{di}$.

    \subsection{Suppression Weight Generation Stage}
        
        Even if easy distractors are excluded through the previous stage, untrimmed videos still contain hard distractors that cannot be easily distinguished and are unrelated to the overall topic of the video due to the various content entanglements. In this section, the Temporal Saliency Module~(TSM) and Topic Guidance Module~(TGM) are introduced for calculating suppression weights, which indicate how close the remaining frames are to the hard distractor. TSM assesses the significance of each frame based on saliency information derived from frame-level similarities, while TGM measures the degree to which each frame relates to the overall topic of the video. The weights obtained from these two modules are converted into the suppression weights~$W$ using the Hadamard product.
        
        \paragraph{Temporal Saliency Module \\} 
            To measure the importance of each frame, saliency information is extracted in the training phase. This is inspired by ViSiL~\cite{kordopatis2019visil}, a model that refines a frame-level similarity map during training and accumulates it to a frame-level similarity via the Chamfer Similarity~(CS)~\cite{barrow1977parametric} operation. Specifically, as the model is optimized, the CS operation leads to an increase in locations, which helps improve video-level similarity within a similarity map of a positive pair. Because of this, the increased locations contain the frames with a strong correlation between the positive pair (as proven in the supplementary material). Therefore, we propose a modified structure that can exploit this correlation as saliency information in TSM by extracting pseudo-labels based on these locations.
        
\begin{figure}[t!]
                \centering
                \includegraphics[width=0.9\linewidth]{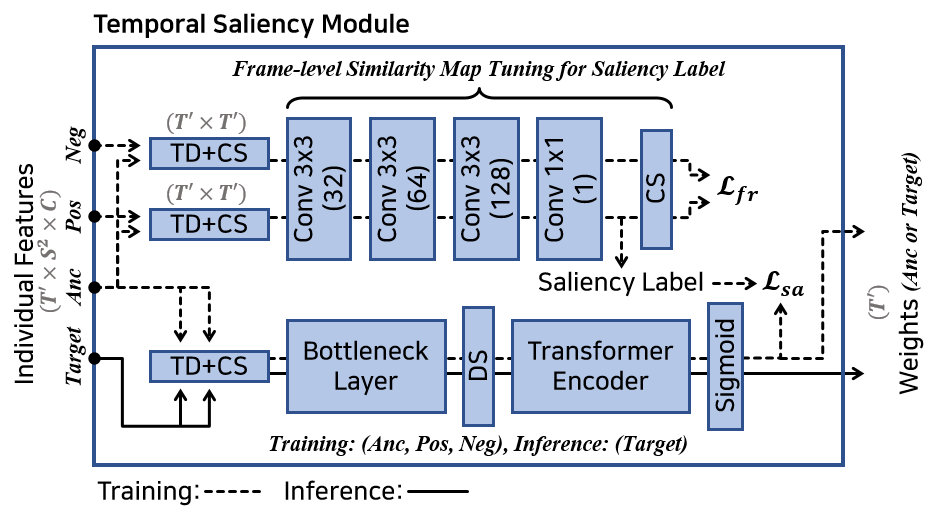} 
                \caption{\textbf{Pipeline of TSM.} The gray italic letters represent the size of the feature in each process. The number in parentheses in the layer blocks indicates the output dimension.} \label{fig:tsm}
            \end{figure}

            Technically, as shown in~\Cref{fig:tsm}, frame-level features of the triplet are transformed by Tensor Dot (TD) and CS into a similarity map for the positive pair~(i.e., anchor and positive) and a similarity map for the negative pair~(i.e., anchor and negative). These similarity maps are then converted into tuned similarity maps~$\mathcal{D}_p$ and $\mathcal{D}_n$ for the positive pair and the negative pair, respectively, through four convolutional layers. Here, we generate a pseudo-label~$Y_{sa}$  (i.e., saliency label) based on the increasing value within~$\mathcal{D}_p$ in order to extract saliency information. This is formulated in~\Cref{slabel_extract}, where the superscript $\textup{\textbf{T}}$ is the transpose operation, and $H$ is the Heaviside step function. Furthermore, $\rho$ is the highest similarity of each frame in the anchor video for the positive video. The saliency label consists of values where $\rho_{i}$ is $1$ if it is greater than the average of $\rho$ and $0$ if it is less, thereby labeling the frame locations that indicate a strong correlation between the positive pair.

            {\small
                \begin{align}
                    \rho_{i} &= \max_{j\in [1,T^{\prime \prime }]} \mathcal{D}_{p}^{(i,j)}, \nonumber
                    \\
                    \rho &= \left [\: \rho_1, \; \rho_2, \; \cdots, \; \rho_{i}, \; \cdots, \; \rho_{T''} \right ]^{\textup{\textbf{T}}},
                    \nonumber
                    \\
                    Y_{sa} &= H ( \rho - \frac{1}{T^{\prime \prime }}\sum_{i=1}^{T^{\prime \prime }} \rho_i ) \in \mathbb{R}^{T^{\prime \prime }}.\label{slabel_extract}
                \end{align}
            }%

            After completing the procedure for creating the saliency label, a self-similarity map is generated by applying TD and CS to two inputs consisting solely of the anchor. The self-similarity map is subsequently fed into the bottleneck layer, the transformer encoder, and the sigmoid to yield saliency weights~$W_{sa}$~$=$~$\{ w^{(t)}_{sa} \}^{T'}_{t=1}$, as shown in~\Cref{fig:tsm}. Here, only diagonal components are sampled from the output map of the previous layer to match the input format when entered into the transformer encoder, i.e., Diagonal Sampling~(DS). Consequently, to enable TSM to recognize salient frames through training, the saliency loss~$\mathcal{L}_{sa}$ is computed as the binary cross entropy loss between the saliency weights $W_{sa}$ and the saliency label $Y_{sa}$, where the nearest interpolation is applied to the label to match the length of the output,~$T'$. The saliency loss~$\mathcal{L}_{sa}$ is optimized with the frame loss~$\mathcal{L}_{fr}$ for tuning the similarity map of the positive pair, which is covered in detail in the supplementary material. During the inference phase, only a self-similarity map for a given target video is fed into the layers to yield saliency weights.

        \paragraph{Topic Guidance Module \\}
        
            The topic of the video is also one of the factors that determines the importance of frames. For this reason, we create an initial state $I$ that gives direct, video-specific instruction on the topic to help the model generate guidance weights~$W_{gu}$~$=$~$\{ w^{(t)}_{gu} \}^{T'}_{t=1}$. More specifically, a rough topic representation~$G$ is initially constructed to roughly represent the topic of the video. According to the claim~\cite{lin2017hnip} that statistical moments (i.e., mean, max, etc.) have been mathematically proven to be invariant across multiple transformations, the ST-GAP, which consists of average operations, is used to create a $G \in \mathbb{R}^{C}$ that is robust to specific transformations between the frame-level features~$X$. In fact, the topic of a video (even if untrimmed) is determined by what most of the content in that video represents. Therefore, since the average operation yields the direction in which most of the content vectors (i.e., frame-level features) point, an approximate (even if simple) representation of the topic can be obtained. As a result, the cosine similarity between $G$ and $X$ is employed to build the initial state $I \in \mathbb{R}^{T'}$, which guides the model to reference the topic. At this time, for convenience of operation, the S-GAP is applied to the frame-level features~$X$ to remove its spatial axis.

\begin{figure}[t!]
            \centering
            \includegraphics[width=0.93\linewidth]{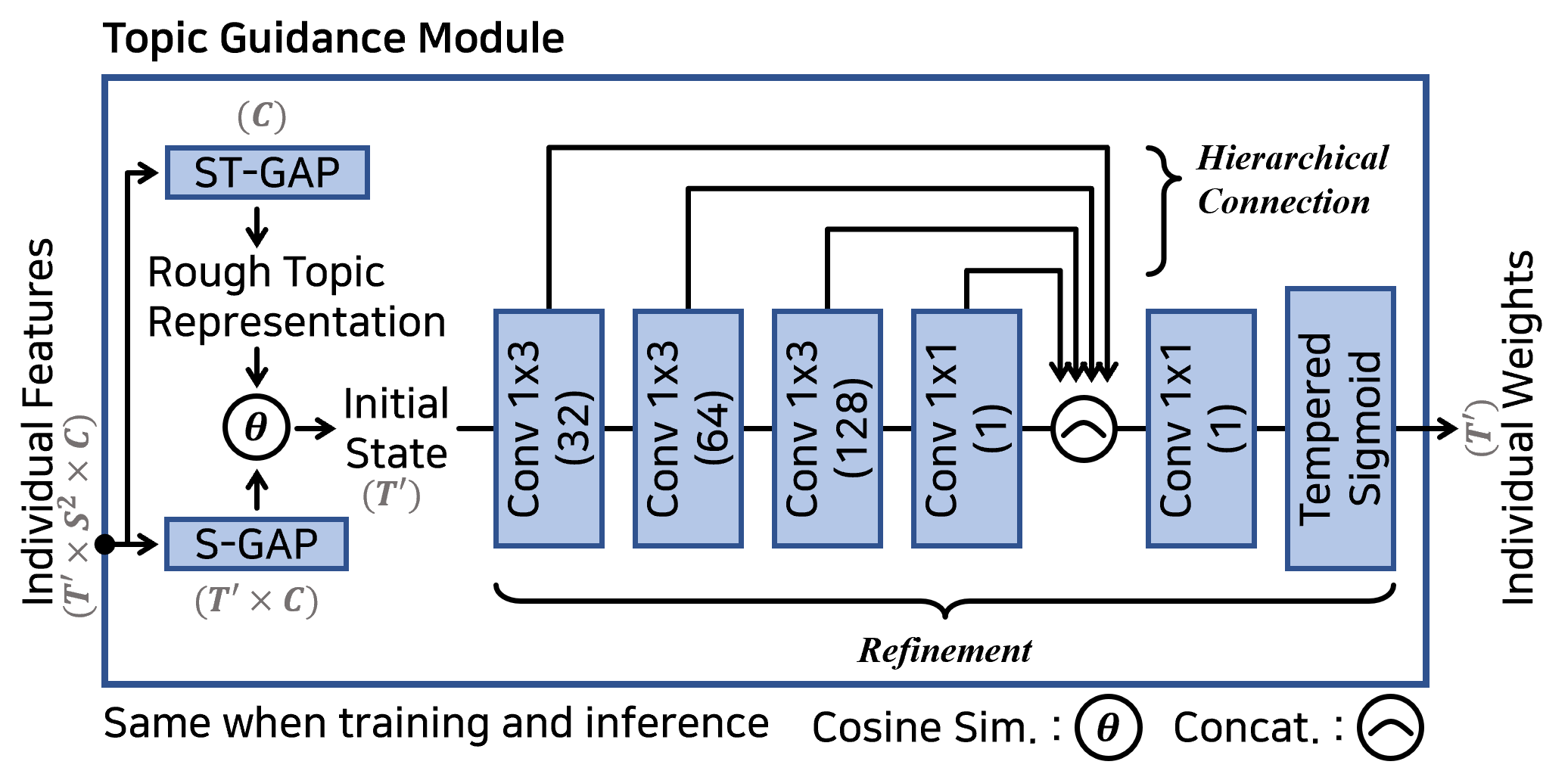} 
            \caption{\textbf{Pipeline of TGM.} The gray italic letters represent the size of the feature in each process. The number in parentheses in layer blocks indicates the output dimension.} \label{fig:tgm}
        \end{figure}

            The initial state~$I$ is effective in directing the model in a rough pattern along the optimal path to the goal; however, a process of refinement must be added with the purpose of providing the guidance weights that more precisely suggest topic relevance. Thus, as illustrated in~\Cref{fig:tgm}, architecture is designed to refine the coarse pattern. With the initial state~$I$ of length~$T'$ as input, the data is collected by sliding 1$\times$3~kernels in three convolutional layers, and then a~1$\times$1~convolutional layer reduces the channel dimension. As the preceding three layers are traversed, the receptive field expands, indicating that the temporal spans of data gathered by these layers extend from the short-term to the long-term. Therefore, the output of the preceding three layers and the output of the 1$\times$1~convolutional layer is designed to channel-wise concatenate, which is referred to as a hierarchical connection, to assist the model in grasping the topic relevance of each frame through direct utilization of the knowledge over various temporal spans. Then, a convolutional layer is applied to shrink the dimension of the channel axis. Only this module employs the tempered sigmoid proposed by~\cite{papernot2021tempered} rather than the sigmoid to reliably learn the weights from noises that may arise during the refining operation from rough patterns.

    \subsection{Video Embedding \& Training Strategy}
        
        In the training phase, frame-level features are aggregated into a video-level feature~$V$~$\in $~$\mathbb{R}^{C}$ by the Hadamard product with the suppression weights $W$ calculated for each video in a triplet: an anchor, a positive, and a negative. At this time, in the case of positive and negative, only $W_{gu}$ is used as the suppression weights $W$ because their weights are not handled in TSM. As a result, the video loss~$\mathcal{L}_{vi}$ is computed as the triplet margin loss between the three video-level features in the triplet. This loss, along with the three losses discussed above, optimize the model according to~\Cref{eq:loss} as,
        \begin{equation}
            \small
            \mathcal{L} = \mathcal{L}_{vi} + \mathcal{L}_{fr} + \mathcal{L}_{sa} + \alpha\mathcal{L}_{di}.
            \label{eq:loss}
        \end{equation}
        In addition, our approach follows the mining scheme of~\cite{kordopatis2019visil} for videos consisting of triplets. $\alpha$ is a parameter for adjusting the learning of DDM as it is faster than other modules when observed empirically. Due to space limitations, further details can be found in the supplementary material.


\section{Experiments}\label{exp}
    \subsection{Evaluation Setup}
        Our experiments were evaluated on two retrieval settings\footnote{Some videos from EVVE~\cite{revaud2013event}, a dataset for event video retrieval (EVR), another common evaluation setting, could not be downloaded. However, for further comparison, the benchmark for a subset we own~($\approx$70.5\% of the original) is covered in the supplementary material.} that are now widely used in CBVR: fine-grained incident video retrieval (FIVR) and near-duplicate video retrieval (NDVR). All performance evaluations are reported based on the mean average precision~(mAP)~\cite{zhu2004recall}, and the implementation details are covered in the supplementary material. Furthermore, VCDB~\cite{jiang2014vcdb} was used as a training dataset, and FIVR~\cite{kordopatis2019fivr} and CC\_WEB\_VIDEO~\cite{wu2009real} were used as evaluation datasets.
        
        \textbf{VCDB} is aimed at video copy detection and consists of 528 core datasets with 9,236 partially copied pairs and about 100,000 videos with no additional metadata. 

        \textbf{FIVR} is equivalent to the FIVR task, which seeks videos connected to certain disasters, occurrences, and incidents. Furthermore, depending on the level of relevance desired, it is evaluated using three criteria: duplicate scene video retrieval (DSVR), complementary scene video retrieval (CSVR), and incident scene video retrieval (ISVR). In this dataset, there are two types in the family: FIVR-5K and FIVR-200K. FIVR-5K has 50 queries and 5,000 videos in the database, while the FIVR-200K has 100 queries and 225,960 videos in the database, both of which have video-level annotations. FIVR-5K is a subset of the FIVR-200K used for ablation studies, and FIVR-200K is used for benchmarking as a large-scale video collection.
        
        \textbf{CC\_WEB\_VIDEO} corresponds to the NDVR task, which aims to find geometrically or photometrically transformed videos. It consists of 13,129 videos in a set of 24 queries and has two types of criteria for evaluation which are divided into evaluations within each query set or within the entire video, and with the original annotation or the ``cleaned" version of the annotation by \cite{kordopatis2019visil}. The combination of these criteria provides four evaluations.

        \begin{table}[!t] 
        \begin{center}
        \small
        \begin{tabular}{@{}clrccc@{}}
            \cmidrule[\heavyrulewidth]{1-6}
            \morecmidrules
            \cmidrule[\heavyrulewidth]{1-6} 
            \multirow{2}{*}{} & \multicolumn{1}{c}{\multirow{2}{*}[-.3em]{Approach}} & \multicolumn{1}{c}{\multirow{2}{*}[-.3em]{\textit{Dim.}}} &
            \multicolumn{3}{c}{FIVR-200K} \\ \cmidrule(l){4-6} 
            &  &  & DSVR & CSVR & \multicolumn{1}{c}{ISVR} \\ \midrule
            \multicolumn{1}{c}{\multirow{8}{*}{\rotatebox[origin=c]{90}{\textit{\textbf{frame}}}}} & TN & \multicolumn{1}{r}{-}\,\, & 0.724 & 0.699 & \multicolumn{1}{c}{0.589}\\
            \multicolumn{1}{l}{} & DP & \multicolumn{1}{r}{-}\,\,  & 0.775 & 0.740 & \multicolumn{1}{c}{0.632} \\
            \multicolumn{1}{l}{} & $\textup{TCA}_{sym}$  & 1,024 & 0.728 & 0.698 & \multicolumn{1}{c}{0.592}  \\
            \multicolumn{1}{l}{} & $\textup{TCA}_{f}$  & 1,024 & 0.877 & 0.830 & \multicolumn{1}{c}{\textbf{0.703}} \\
            \multicolumn{1}{l}{} & TNIP  & 1,040 & \textbf{0.896} & 0.833 & \multicolumn{1}{c}{0.674} \\ 
            \multicolumn{1}{l}{} & $\textup{ViSiL}_{sym}$  & 3,840 & 0.833 & 0.792 & \multicolumn{1}{c}{0.654}  \\
            \multicolumn{1}{l}{} & $\textup{ViSiL}_{f}$ & 3,840 & 0.843 & 0.797 & \multicolumn{1}{c}{0.660}  \\
            \multicolumn{1}{l}{} & $\textup{ViSiL}_{v}$  & 3,840 & 0.892 & \textbf{0.841} & \multicolumn{1}{c}{0.702} \\ \midrule 
            \multicolumn{1}{c}{\multirow{10}{*}[0em]{\rotatebox[origin=c]{90}{\textit{\textbf{video}}}}} & HC  & \multicolumn{1}{r}{-}\,\, & 0.265 & 0.247 & \multicolumn{1}{c}{0.193}\\
             & DML  & 500 & 0.398 & 0.378 & \multicolumn{1}{c}{0.309}  \\ 
             & TMK  & 65,536 & 0.417 & 0.394 & \multicolumn{1}{c}{0.319} \\ 
             & LAMV  & 65,536 & 0.489 & 0.459 & \multicolumn{1}{c}{0.364} \\
             & VRAG  & 4,096 & 0.484 & 0.470 & \multicolumn{1}{c}{0.399} \\
             & $\textup{TCA}_{c}$  & 1,024 & 0.570 & 0.553 & \multicolumn{1}{c}{0.473} \\ \cmidrule(l){2-6}
             & $\textbf{\textup{VVS}}_{\textit{500}}$ \,\,\textbf{(ours)} & 500 & 0.606 & 0.588 & \multicolumn{1}{c}{0.502} \\
             & $\textbf{\textup{VVS}}_{\textit{512}}$ \,\,\textbf{(ours)} & 512 & 0.608 & 0.590 & \multicolumn{1}{c}{0.505} \\
             & $\textbf{\textup{VVS}}_{\textit{1024}}$ \textbf{(ours)} & 1,024 & 0.645 & 0.627 & \multicolumn{1}{c}{0.536} \\
             & $\textbf{\textup{VVS}}_{\textit{3840}}$ \textbf{(ours)} & 3,840 & \textbf{0.711} & \textbf{0.689} & \multicolumn{1}{c}{\textbf{0.590}} \\
            \cmidrule[\heavyrulewidth]{1-6}
            \morecmidrules
            \cmidrule[\heavyrulewidth]{1-6} 
            \end{tabular} 
        \caption{\textbf{Benchmark on FIVR-200K.} The \textit{\textbf{frame}} and \textit{\textbf{video}} refer to frame-level and video-level feature-based approaches. \textit{Dim.}~refers to the dimension of the basic unit for calculating similarity in each approach (i.e., frame-level approaches use multiple features of that dimension, as many as the number of all or most frames in a video, while video-level approaches use only one feature of that dimension). Only approaches that are trained from VCDB or do not require additional training are shown for a fair comparison.} \label{tab:fivr}
        \end{center}
    \end{table}
    
    \begin{table}[t] 
        \begin{center}
        \small
        \setlength{\tabcolsep}{4.5pt}
        \begin{tabular}{clrcccc}
            \cmidrule[\heavyrulewidth]{1-7}
            \morecmidrules
            \cmidrule[\heavyrulewidth]{1-7} 
            \multirow{2}{*}{} & \multicolumn{1}{c}{\multirow{2}{*}[-.3em]{Approach}} & \multicolumn{1}{c}{\multirow{2}{*}[-.3em]{\textit{Dim.}}} &
            \multicolumn{4}{c}{CC\_WEB\_VIDEO} \\ \cmidrule(){4-7} 
            & & & \multicolumn{1}{c}{\small{$\textup{cc}$}} & \multicolumn{1}{c}{\small{$\textup{\:cc}^{*}$}} & \multicolumn{1}{c}{\small{$\textup{cc}_{c}$}} & \multicolumn{1}{c}{\small{$\textup{cc}_{c}^{*}$}} \\ \midrule     
            \multicolumn{1}{c}{\multirow{9}{*}{\rotatebox[origin=c]{90}{\textit{\textbf{frame}}}}} & TN & \multicolumn{1}{r}{-}\,\, & 0.978 & 0.965 & 0.991 & 0.987 \\
            \multicolumn{1}{l}{} & DP & \multicolumn{1}{r}{-}\,\, & 0.975 & 0.958 & 0.990 & 0.982 \\
            \multicolumn{1}{l}{} & CTE & \multicolumn{1}{r}{-}\,\, & \textbf{0.996} & \multicolumn{1}{c}{-} & \multicolumn{1}{c}{-} & \multicolumn{1}{c}{-} \\
            \multicolumn{1}{l}{} & $\textup{TCA}_{sym}$  & 1,024 & 0.982 & 0.962 & 0.992 & 0.981 \\
            \multicolumn{1}{l}{} & $\textup{TCA}_{f}$ & 1,024 & 0.983 & 0.969 & 0.994 & 0.990 \\
            \multicolumn{1}{l}{} & TNIP  &  \multicolumn{1}{r}{1,040} & 0.978 & 0.969 & 0.983 & 0.975 \\ 
            \multicolumn{1}{l}{} & $\textup{ViSiL}_{sym}$  & 3,840 & 0.982 & 0.969 & 0.991 & 0.988 \\
            \multicolumn{1}{l}{} & $\textup{ViSiL}_{f}$  & 3,840 & 0.984 & 0.969 & 0.993 & 0.987 \\
            \multicolumn{1}{l}{} & $\textup{ViSiL}_{v}$  & 3,840 & 0.985 &\textbf{0.971} & \textbf{0.996} & \textbf{0.993} \\ \midrule
            \multicolumn{1}{c}{\multirow{8}{*}{\rotatebox[origin=c]{90}{\textit{\textbf{video}}}}} & HC  & \multicolumn{1}{r}{-}\,\, & 0.958 & \multicolumn{1}{c}{-} & \multicolumn{1}{c}{-} & \multicolumn{1}{c}{-} \\ 
             & DML  & \multicolumn{1}{r}{500} & 0.971 & 0.941 & 0.979 & 0.959 \\ 
             & VRAG  & \multicolumn{1}{r}{4,096} & 0.971 & 0.952 & 0.980 & 0.967 \\
             & $\textup{TCA}_{c}$ & 1,024 & 0.973 & 0.947 & 0.983 & 0.965 \\ \cmidrule(l){2-7} 
             & $\textbf{\textup{VVS}}_{\textit{500}}$\,\,\,\textbf{(ours)}  & 500  & 0.973 & 0.952 & 0.981 & 0.966 \\
             & $\textbf{\textup{VVS}}_{\textit{512}}$\,\,\,\textbf{(ours)}  & 512  & 0.973 & 0.952 & 0.981 & 0.967 \\
             & $\textbf{\textup{VVS}}_{\textit{1024}}$\,\textbf{(ours)}  & 1,024 & 0.973 & 0.952 & 0.982 & 0.969 \\
             & $\textbf{\textup{VVS}}_{\textit{3840}}$\,\textbf{(ours)}  & 3,840 & \textbf{0.975} & \textbf{0.955} & \textbf{0.984} & \textbf{0.973} \\
            \cmidrule[\heavyrulewidth]{1-7}
            \morecmidrules
            \cmidrule[\heavyrulewidth]{1-7} 
        \end{tabular} 
        \caption{\textbf{Benchmark on CC\_WEB\_VIDEO.} (*) refers to the evaluation of the entire dataset, and the subscript $c$ refers to the use of cleaned annotations. All other notations and settings are identical to those presented in \Cref{tab:fivr}.} \label{tab:ccweb}
        \end{center}     
    \end{table} 
        \begin{table}[!t]
            \small
            \centering
            \begin{tabular}{c|ccc|ccc}
                \cmidrule[\heavyrulewidth]{1-7}
                \morecmidrules
                \cmidrule[\heavyrulewidth]{1-7}
                & \multicolumn{1}{c|}{\textit{Elim.}}  & \multicolumn{2}{c|}{\textit{Gen.}} & \multicolumn{3}{c}{FIVR-5K} \\ \cmidrule(){2-7} 
               
                    &\multicolumn{1}{c|}{DDM} & TSM    & TGM    & DSVR  & CSVR  & ISVR  \\ \midrule 
                (a) &       &        &        & 0.692 & 0.700 & 0.651 \\ \midrule
                (b) &\cmark &        &        & 0.715 & 0.725 & 0.672 \\ \midrule
                (c) &       & \cmark &        & 0.702 & 0.710 & 0.661 \\
                (d) &       &        & \cmark & 0.716 & 0.724 & 0.677 \\
                (e) &       & \cmark & \cmark & 0.719 & 0.726 & 0.680 \\ \midrule 
                (f) &\cmark & \cmark &        & 0.724 & 0.732 & 0.683 \\ 
                (g) &\cmark &        & \cmark & 0.738 & 0.746 & 0.698 \\ \midrule
                (h) &\cmark & \cmark & \cmark & \textbf{0.744} & \textbf{0.752} & \textbf{0.705} \\
                \cmidrule[\heavyrulewidth]{1-7}
                \morecmidrules
                \cmidrule[\heavyrulewidth]{1-7}
            \end{tabular}
            \caption{\textbf{Module-wise Ablations for $\textbf{\textup{VVS}}_{\textit{3840}}$\textbf{.} }\textit{Elim.} refers to the easy distractor elimination stage, and \textit{Gen.} refers to the suppression weight generation stage. (a) represents a baseline of the same dimension that weighs all frames equally without any of the proposed modules, (b)-(g) represent module-wise ablations, and (h) represents $\textup{VVS}_{\textit{3840}}$.}
            \label{tab:module_ablation}
        \end{table}

    \subsection{Comparison with Other Approaches}

        Based on the dimension~$C$ of a video-level feature $V$, the proposed approach is referred to as $\textup{VVS}_C$. $C$ is equal to that of a frame-level feature~$X$ and is determined by dimension reduction during the PCA whitening procedure. If dimension reduction is not applied, it is $\textup{VVS}_{\textit{3840}}$ (as used in~\cite{kordopatis2019visil}, the dimension of $\textup{L}_{N}\textup{-iMAC}$ is 3840), and if dimension reduction is applied to match the dimension with other approaches, it is $\textup{VVS}_{\textit{500}}$, $\textup{VVS}_{\textit{512}}$ and $\textup{VVS}_{\textit{1024}}$.

        \Cref{tab:fivr} shows comparisons with previous state-of-the-art approaches on the large-scale FIVR-200K dataset. In this dataset, $\textup{VVS}_{\textit{3840}}$ performs approximately 25\% better than the leading video-level approach in all tasks, which is close to the borderline of the frame-level state-of-the-art approaches. In addition, our approaches~$\textup{VVS}_{\textit{500}}$, $\textup{VVS}_{\textit{512}}$ and $\textup{VVS}_{\textit{1024}}$ are state-of-the-art regardless of whether their dimensions match or are smaller than those of other video-level approaches. This trend is similar to the performance on the CC\_WEB\_VIDEO in~\Cref{tab:ccweb}. This proves that our method is the most optimal framework between the two streams, considering that video-level approaches are essentially memory- and speed-efficient.

    \begin{table}[!t]
    \small
    \centering
    \begin{tabular}{cc|ccc}
        \cmidrule[\heavyrulewidth]{1-5}
        \morecmidrules
        \cmidrule[\heavyrulewidth]{1-5} 
        \multirow{6}{*}[-.5em]{DDM} & \multirow{2}{*}[-.3em]{\begin{tabular}[c]{@{}c@{}}Injection\\ Ratio\end{tabular}} & \multicolumn{3}{c}{FIVR-5K} \\ \cmidrule(){3-5} 
         &  & DSVR & CSVR & ISVR \\ \cmidrule(){2-5} 
         & \multicolumn{1}{c|}{\:\;0\% -\:20\%} & 0.739 & 0.748 & 0.701 \\
         & \multicolumn{1}{c|}{20\% -\:50\%}    & \textbf{0.744} & \textbf{0.752} & \textbf{0.705} \\
         & \multicolumn{1}{c|}{50\% -\:80\%}    & 0.738 & 0.749 & 0.704 \\ 
         & \multicolumn{1}{c|}{80\% -\:100\%}   & 0.728 & 0.743 & 0.701 \\
         \cmidrule[\heavyrulewidth]{1-5}
        \morecmidrules
        \cmidrule[\heavyrulewidth]{1-5}
    \end{tabular}
    \caption{\textbf{Distractor Injection Ratio in DDM.} This demonstrates the overall impact according to the sampling ratio of the easy distractor set in DDM.}
    \label{tab:compo_ablation_ddm}
    \end{table}

    \begin{table}[!t]
    \small
    \centering
    \begin{tabular}{cc|ccc}
        \cmidrule[\heavyrulewidth]{1-5}
        \morecmidrules
        \cmidrule[\heavyrulewidth]{1-5}  
        \multirow{4}{*}[-0.6em]{TSM} & \multirow{2}{*}[-.3em]{\begin{tabular}[c]{@{}c@{}}Frame\\ Loss $\mathcal{L}_{fr}$\end{tabular}} & \multicolumn{3}{c}{FIVR-5K} \\ \cmidrule(){3-5}
         &  & DSVR  & CSVR  & ISVR  \\ \cmidrule(){2-5} 
         &  & 0.742 & 0.749 & 0.702 \\
         & \cmark & \textbf{0.744} & \textbf{0.752} & \textbf{0.705} \\
         \cmidrule[\heavyrulewidth]{1-5}
        \morecmidrules
        \cmidrule[\heavyrulewidth]{1-5}
    \end{tabular}
    \caption{\textbf{Existence of Frame Loss $\mathcal{L}_{fr}$ in TSM.} This demonstrates how frame loss affects TSM.}
    \label{tab:compo_ablation_tsm}
    \end{table}
        
    \begin{table}[!t]
    \small
    \centering
    \begin{tabular}{ccc|ccc}
        \cmidrule[\heavyrulewidth]{1-6}
        \morecmidrules
        \cmidrule[\heavyrulewidth]{1-6} 
        \multicolumn{3}{c|}{TGM}                        & \multicolumn{3}{c}{FIVR-5K} \\ \cmidrule(){1-6}
        \textit{Init.} & \textit{Refine.} & \textit{Hier.} & DSVR    & CSVR    & ISVR    \\ \midrule
        \multicolumn{1}{l}{\textit{Rand.}}       & \cmark               & \cmark   & 0.693   & 0.701   & 0.652   \\
        \multicolumn{1}{l}{\textit{Const.}}      & \cmark               & \cmark   & 0.693   & 0.701   & 0.652   \\ \midrule
        \multicolumn{1}{l}{$G$}                  &                      &          & 0.625   & 0.631   & 0.584   \\
        \multicolumn{1}{l}{$G$}                  & \cmark               &          & 0.712   & 0.722   & 0.675   \\
        \multicolumn{1}{l}{$G$}                  & \cmark               & \cmark   & \textbf{0.716} & \textbf{0.724} & \textbf{0.677}   \\
        \cmidrule[\heavyrulewidth]{1-6}
        \morecmidrules
        \cmidrule[\heavyrulewidth]{1-6} 
    \end{tabular}
    \caption{\textbf{Structure within TGM.} This demonstrates the impact of the structure within TGM. \textit{Init.} refers to the initial state $I$, \textit{Refine.} to the refinement process, and \textit{Hier.} to the hierarchical connection. \textit{Rand.} and \textit{Const.} refer to situations in which the initial state is formed from a random or constant value (which is~0.5), not the rough topic representation $G$. To facilitate independent evaluation, the framework excludes all modules except TGM.}
    \label{tab:compo_ablation_tgm}
    \end{table}

    \begin{figure}[t!]
        \centering
        \includegraphics[width=0.9\linewidth]{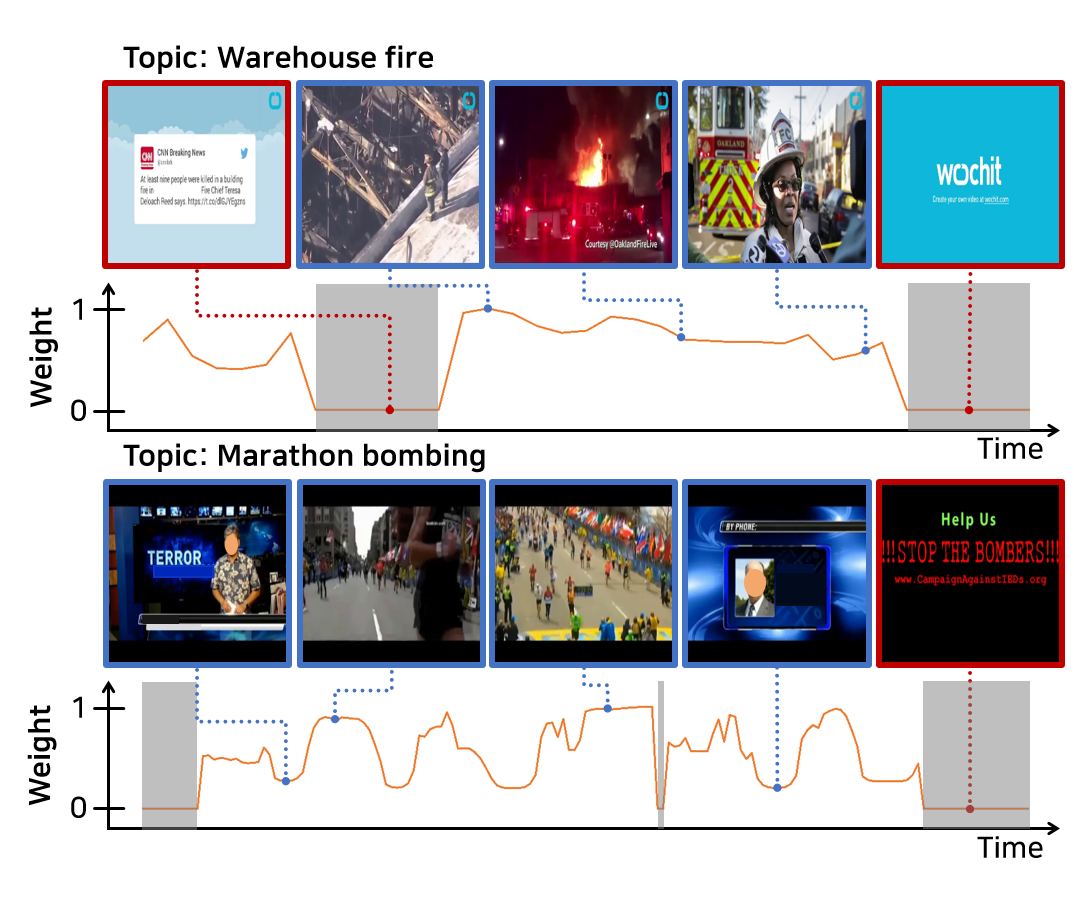} 
        \caption{\textbf{Qualitative Results on FIVR-5K.} The orange line refers to the weights from TSM and TGM; the lower the value, the more suppressed the frame. The gray region corresponds to easy distractors eliminated by DDM, and frames that belong to this area are denoted by a red border.}
        \label{fig:qual}
    \end{figure}
    
    \subsection{Ablation Studies \& Analyses}

        \paragraph{Module-wise Ablations \\}
            This section covers ablation studies for each module in the proposed framework~$\textup{VVS}_{\textit{3840}}$. As seen in~\Cref{tab:module_ablation}, each module~(b)-(d) demonstrates a significant performance increase over the baseline~(a), demonstrating their value. In addition, improvements are observed even when modules are paired with one another~(e)-(g), and the same is true when they are all combined~(h). Moreover, in the supplementary material, by presenting further module-wise ablations of~$\textup{VVS}_{\textit{500}}$, $\textup{VVS}_{\textit{512}}$ and $\textup{VVS}_{\textit{1024}}$, we show that all modules in our approach have a similar impact.

        \paragraph{Component-wise Ablations \\}

            This section covers ablation studies for components within each module of the proposed framework.
            
            \noindent\textit{\textbf{Distractor Injection Ratio in DDM.}}
            \Cref{tab:compo_ablation_ddm} demonstrates the effect of the sampling ratio from the easy distractor set for injection during the training phase in DDM. The model can learn slightly more cases for easy distractors compared to a lower ratio when the input length is 20–50\% relative to $T$, leading to enhancements in the overall framework. However, when selected at a higher ratio, the proportion of frames corresponding to the distractor in a video increases excessively, which hinders optimization.

            \noindent\textit{\textbf{Existence of Frame Loss in TSM.}}
            To assess the impact of frame loss on TSM, \Cref{tab:compo_ablation_tsm} shows the outcomes of ablation when only TSM exists with no other modules. In conclusion, the frame loss allows the saliency information to be tuned, resulting in a more exact saliency label and a boost in performance.

            \noindent\textit{\textbf{Structure within TGM.}}
            To test the validity of the TGM structure, ablation studies for each component are shown in~\Cref{tab:compo_ablation_tgm}, and all modules other than TGM are omitted for independent evaluation of each component. First, if random or constant values are used instead of the rough topic representation~$G$ while constructing the initial state, performance deteriorates, as the model is implicitly required by relatively unclear criteria rather than explicitly guided by the topic to be well optimized. In addition, the performance gap demonstrates that even with~$G$, the refinement process with a hierarchical connection is necessary to direct the model appropriately. Furthermore, as detailed in the supplementary material, the hierarchical connection can make the model more robust for various video lengths.

        \paragraph{Qualitative Results \\}
            \Cref{fig:qual} depicts the qualitative outcomes produced by the proposed framework. In the first example, where the topic is ``Warehouse fire'', it can be seen that the frames predicted by DDM as the distractors have few low-level characteristics. In addition, the fourth frame in this example is assigned a relatively low weight because no visual clues directly related to the topic appear. In the second example, where the topic is ``Marathon bombing'', it is shown that frames containing only text and low-level characteristics are deleted by DDM, as in the first example. Furthermore, among the remaining frames, the weights of those visually related to the topic are measured to be high, whereas the weights of the first and fourth frames, in which the scene of the event is not shown directly, are low. From these two examples, it is clear that the proposed approach achieves its intended results.
    
\section{Conclusion}
    In this paper, we demonstrate that suppression of irrelevant frames is essential in describing an untrimmed video with long and varied content as a video-level feature. To achieve this, we present an end-to-end framework: VVS. VVS removes frames that can be clearly identified as distractors and determines the degree to which remaining frames should be suppressed based on saliency information and topic relevance. Thus, this approach is the first designed to be learned by explicit criteria, unlike previous approaches that have optimized the model implicitly. Consequently, extensive experiments proved the validity of this design and, at the same time, demonstrated that it is closest to the ideal search scenario among existing approaches due to its competitive speed and efficient memory utilization, as well as its state-of-the-art search accuracy. We hope that this work can contribute to the advancement of real-world video search systems.

\section*{Acknowledgements}
    This work was partly supported by the Institute of Information and Communications Technology Planning and Evaluation (IITP) grant funded by the Korea government (MSIT) (No.2021-0-02067, Next Generation AI for Multi-purpose Video Search, 50\%) and (RS-2023-00231894, Development of an AI Based Video Search System for Untact Content Production, 50\%).

\bibliography{aaai24}

\clearpage
\appendix
\setcounter{figure}{0}
\setcounter{equation}{0}
\setcounter{table}{0}

\renewcommand\thefigure{\Alph{figure}}
\renewcommand\theequation{\roman{equation}}
\renewcommand\thetable{\Alph{table}}

\twocolumn[\section{\begin{LARGE} Supplementary Material \vspace{15mm}\end{LARGE}}]

\setcounter{secnumdepth}{2}

\subparagraph{}
In this supplementary material, we offer extra information to supplement the main script. \Cref{detail} focuses mainly on details. \Cref{imple} offers the implementation details of the proposed framework, and \Cref{strategy} discusses the training strategy details. \Cref{formula_operation} addresses the details of the basic operations to describe the shape change of tensors. \Cref{formula} covers formulas details of the losses to train our approach. \Cref{FDTT} presents additional information on the process of extracting the saliency label while tuning similarity maps. \Cref{arch} provides architecture details via detailed pipelines. \Cref{ablation_analyses} focuses mainly on additional ablation studies and analyses. As stated in the introduction of the main script, \Cref{quan_exp} explains the quantitative experiment indicating that the description of video-level features with optimal suppression of irrelevant frames can be an ideal retrieval scenario. \Cref{ddm_exam} offers examples within the easy distractor set and describes the configuration process of the set. \Cref{training_ddm} explains the reason for training DDM as opposed to simply observing the magnitude of frame-level features to remove easy distractors in the inference phase. \Cref{tempered_exp} discusses the advantages of using the tempered sigmoid rather than the sigmoid in our approach. As stated in the main script while introducing TSM, \Cref{effect_saliency_signal} reveals that saliency information (i.e., saliency signals) can enhance the representation of video-level features. \Cref{hcon_eff} demonstrates why TGM, a module within the proposed framework, includes the refinement process via the hierarchical connection. \Cref{speed_memory} includes numerical comparisons between the proposed approach and other approaches in terms of speed and memory. \Cref{bench_evve} provides an additional benchmark for the subset of the EVVE dataset for the reasons stated in the evaluation setup footnote of the main script. \Cref{applicability} presents the further applicability of the proposed suppression scheme to video summarization. \Cref{further_abl} addresses further module-wise ablations for the proposed approaches:~$\textup{VVS}_{\small{\textit{500}}}$, $\textup{VVS}_{\small{\textit{512}}}$, and $\textup{VVS}_{\small{\textit{1024}}}$. \Cref{qual_exp} contains numerous qualitative results of the proposed framework.

\section{Additional Details}\label{detail}
    \subsection{Implementation Details}\label{imple}
        Each video is sampled at one frame per second. The number of frames per video is set during training with $T$~$=$~$64$ but is varied during inference based on the duration of the video. In the $\textup{L}_{N}\textup{-iMAC}$, the backbone network~$\Phi$ is ResNet50~\cite{he2016deep}, the number of layers~$K$ is 4, the types of region kernel~$N$ are 4. The set of easy distractors used in the training phase is taken from nearly 100,000 metadata-free videos (with the exception of 528 videos in the core datasets) within the VCDB~\cite{jiang2014vcdb}, where each frame-level feature is selected as an easy distractor based on the magnitude threshold~$\lambda_{mag}$~$=$~$40$. The distractor threshold~$\lambda_{di}$ for the thresholding operation on the confidence of DDM during the inference phase is set to $0.5$. The temperature of the tempered sigmoid~\cite{papernot2021tempered} is set to $512$ in TGM. The parameter~$\alpha$ for balancing the discrimination loss is set to~$0.5$, and margins in the frame loss and the video loss are also set to~$0.5$. Additionally, PCA is incrementally trained from all metadata-free videos within VCDB.
        
        All experiment code is implemented with Pytorch on an NVIDIA Tesla V100. To achieve reasonable time comparisons throughout the inference phase, all speed-related experiments are measured synchronously. Given that the CUDA call from Pytorch is fundamentally asynchronous, this synchronous measurement prevents erroneous measurements of overall operating time on the CPU and GPU. In addition, the total inference time includes model operation time, similarity calculation time, and evaluation time after extracting frame-level features, as in~\cite{ng2022vrag}.

        Experiments that are not explicitly indicated (e.g., $\textup{VVS}_{\small{\textit{500}}}$, $\textup{VVS}_{\small{\textit{512}}}$, or $\textup{VVS}_{\small{\textit{1024}}}$) in the main script and supplementary material were all conducted using $\textup{VVS}_{\small{\textit{3840}}}$.

    \subsection{Training Strategy Details}\label{strategy}
        The proposed approach is learned through the Adam optimizer~\cite{kingma2014adam} with one batch, and a fixed learning rate of $2$$\times$$10^{-5}$, during $60$~epochs using 2,000~iterations. The best model during these epochs is chosen using mAP on the FIVR-5K, considered a validation set in this field of study. In addition, videos from the VCDB, including the core dataset and metadata-free videos, are sampled during the training phase to form a triplet.
        
        Furthermore, easy distractors injected into an input of DDM are derived from metadata-free videos. For this reason, only frame-level features chosen from the core dataset, except metadata-free videos, are used as the input of DDM for learning. Specifically, if an input is provided from metadata-free videos, DDM is trained by injecting easy distractors into the input and then creating a pseudo-label based on the position of the injected distractors. However, before the easy distractors are injected, frames included in the easy distractor set may already exist in the input. Therefore, even if specific frames are included in the set of easy distractors, the pseudo-label may state that they are not easy distractors. Consequently, to avoid confusion during model optimization, DDM is trained solely on frame-level features from the core dataset, with the exception of those from metadata-free videos used to generate the easy distractor set. In addition, unlike other modules, DDM takes frame-level features without PCA as input to directly assess to what extent layers of the backbone network activate low-level characteristics.

    \subsection{Operation Details}\label{formula_operation}
        This section addresses the details of the basic operations used to describe the shape change of tensors to help readers understand our approach.
        \paragraph{Tensor Dot \& Chamfer Similarity  \\}
        Tensor Dot~(TD)~\cite{yang2016deep} between two given tensors $\mathcal{A},\mathcal{B}$~$\in$~$\mathbb{R}^{T \times S^2 \times C}$ is defined as a summation for a specific axis. $T$, $S$, and $C$ refer to the temporal, spatial, and channel axes. We only handle the operation on the channel axis; therefore, the output has the size~$\mathbb{R}^{T \times S^2 \times S^2 \times T}$. Chamfer Similarity~(CS)~\cite{barrow1977parametric} is defined as the average of the maximum values in the column axis for a given matrix~$\mathcal{D} \in \mathbb{R}^{N \times M}$, as shown in \Cref{eq1}. If the size of~$\mathcal{D}$ is $\mathbb{R}^{T \times S^2 \times S^2 \times T}$, the output size will be $\mathbb{R}^{T \times T}$, and if it is $\mathbb{R}^{T \times T}$, the output size will be $\mathbb{R}^{1}$.
        \begin{equation}
            CS(\mathcal{D}) = \frac{1}{N}\sum_{i=1}^{N} \max_{j\in [1,M]} \mathcal{D}^{(i,j)}
            \label{eq1}
        \end{equation}

        \paragraph{Spatio-Temporal Global Average Pooling \\}
            Spatio-Temporal Global Average Pooling~(ST-GAP)~\cite{lin2013network} is the global average pooling of the spatial and temporal axes of a given tensor $\mathcal{A}$~$\in$~$\mathbb{R}^{T \times S^2 \times C}$, yielding the output size~$\mathbb{R}^{C}$. Similarly, Spatial Global Average Pooling~($\textup{S-GAP}$) refers to that pooling on the spatial axis alone, with an output of size $\mathbb{R}^{T \times C}$ for the same input. After these two operations, L2 normalization is applied to adjust the difference in scale between tensors.

        \paragraph{Diagonal Sampling \\}
            Diagonal Sampling (DS)~\cite{kang2022uboco} means extracting the diagonal components of a given square matrix $\mathcal{E}$~$\in$~$\mathbb{R}^{T \times T \times C}$. The input we handle with this operation always consists of three axes, and an output of size $\mathbb{R}^{T \times C}$ is generated by sampling from the other two axes besides the channel axis.

    \subsection{Loss Details}\label{formula}
        This section covers the details of the formulas for each loss, which are omitted in the main script due to their already well-known forms, to help readers understand our approach.
        \paragraph{Discrimination Loss \\}
        The discrimination loss $\mathcal{L}_{di}$ is the form of the binary cross entropy loss between the confidence~$W_{di}$~$=$~$\{ w^{(t)}_{di} \}^{T'}_{t=1}$ and the pseudo-label~$Y_{di}$~$=$~$\{ y^{(t)}_{di}\}^{T'}_{t=1}$ obtained from DDM to identify easy distractors, which can be formulated as follows:
        \begin{equation}
            \mathcal{L}_{di}=\frac{1}{T^{\prime }} \sum^{T^{\prime }}_{t=1} y^{(t)}_{di}\log(w^{(t)}_{di})+(1-y^{(t)}_{di})\log( 1-w^{(t)}_{di}).
        \end{equation}

        \paragraph{Frame Loss \\}
        The frame loss~$\mathcal{L}_{fr}$, which calibrates saliency information in TSM, is the sum of the triplet margin loss for tuned similarity maps and the regularization loss. The triplet margin loss (denoted by $\mathcal{L}_{tri}$) is the form of the well-known triplet margin loss between the tuned similarity maps $\mathcal{D}_{p}$ and $\mathcal{D}_{n}$ for the positive and negative pair calculated within TSM, which can be formulated as follows:
        \begin{gather}
            \mathcal{L}_{tri} =\max \left\{ 0,CS\left( \mathcal{D}_{n}\right) -CS\left(\mathcal{D}_{p}\right) +\gamma \right\},  \label{triplet_loss}
        \end{gather}
        where $\gamma$ is the margin. The regularization loss (denoted by $\mathcal{L}_{reg}$) is a form of divergence constraint between the two tuned similarity maps $\mathcal{D}_{p}$ and $\mathcal{D}_{n}$, which can be formulated as follows:
        \begin{gather} 
            \psi (\mathcal{D}) = \sum{\left| \max \left\{ 0,\mathcal{D}-J\right\} \right| +\left| \min \left\{ 0,\mathcal{D}+J\right\} \right|}, \nonumber
            \\
            \mathcal{L}_{reg} = \psi (\mathcal{D}_{p}) + \psi (\mathcal{D}_{n}), \label{reg_loss}
        \end{gather}
        where $0$ and $J$ are matrices of zeros and ones with the same size as $\mathcal{D}_{p}$ and $\mathcal{D}_{n}$. Consequently, the frame loss~$\mathcal{L}_{fr}$ is the form of the weighted sum between the two prior losses, which can be formulated as follows:
        \begin{equation}
            \mathcal{L}_{fr} = \mathcal{L}_{tri} + 0.5 * \mathcal{L}_{reg}. \label{frame_loss}
        \end{equation}

        \paragraph{Saliency Loss \\}
        The saliency loss $\mathcal{L}_{sa}$ is the form of the binary cross entropy loss between the saliency weights~$W_{sa}$~$=$~$\{ w^{(t)}_{sa} \}^{T'}_{t=1}$ and the saliency label~$Y_{sa}$~$=$~$\{ y^{(t)}_{sa} \}^{T'}_{t=1}$ obtained from TSM to assess the significance of each frame, which can be formulated as follows:
        \begin{equation} \label{saliency_loss}
            \mathcal{L}_{sa}=\frac{1}{T^{\prime }} \sum^{T^{\prime }}_{t=1} y^{(t)}_{sa}\log(w^{(t)}_{sa})+(1-y^{(t)}_{sa})\log (1-w^{(t)}_{sa}).
        \end{equation}

        \paragraph{Video Loss \\}
        The video loss $\mathcal{L}_{vi}$ is the form of the triplet margin loss between video-level features within a triplet, which can be formulated as follows:
        \begin{equation}
            \mathcal{L}_{vi}=\max \left\{ 0,\theta \left( V,V^{-}\right) -\theta \left( V,V^{+}\right) +\gamma \right\},
        \end{equation}
        where $V$, $V^{+}$, and $V^{-}$ are video-level features derived from an anchor, a positive, and a negative within the triplet; $\theta (\cdot )$ is the cosine similarity; and $\gamma$ is the margin.

        Since our approach is simultaneously optimized from all the losses described above, we term it the ``end-to-end framework'' to avoid the misunderstanding that each stage is learned individually (in practice, two stages are trained at once); however, note that the backbone network for extracting $\textup{L}_{N}\textup{-iMAC}$ remains frozen, as is usual in this line of research.

    \subsection{Further Details of Training TSM}\label{FDTT}
        This section provides additional information on the process of extracting the saliency label~$Y_{sa}$ while tuning similarity maps of a positive pair to optimize $\mathcal{L}_{sa}$ and $\mathcal{L}_{fr}$ jointly in TSM during the training phase. The process can be observed by referring to~\Cref{alg:algorithm}. The saliency label is calculated as $\mathcal{L}_{sa}$ using~\Cref{saliency_loss}, and then $\mathcal{L}_{sa}$ and $\mathcal{L}_{fr}$ are minimized at once, as stated in the main script.

    \begin{algorithm}[tb]
        \caption{Process of Extracting Labels and Tuning Maps}
        \label{alg:algorithm}
        \textbf{Input}: Frame-level features $X_{a},X_{p},X_{n}$ in a triplet
        \\
        \null\hfill (i.e., the outputs of DDM with dimension $\mathbb{R}^{T'\times S^2 \times C}$)
        \\
        \textbf{Parameter}: Tuning layers~$f$ parameterized by~$\theta$
        \\
        \begin{algorithmic}[1] 
        \STATE {\# Tune the frame-level similarity maps}
        \STATE $\mathcal{D}_{p}=f(\textup{CS}(\textup{TD}(X_{a}, X_{p}));\theta)\in \mathbb{R}^{T''\times T''}$
        \STATE $\mathcal{D}_{n}=f(\textup{CS}(\textup{TD}(X_{a}, X_{n}));\theta)\in \mathbb{R}^{T''\times T''}$
        \STATE {\# Extract the saliency label}
        \STATE $Y_{sa}=H\left( \rho -\frac{1}{T^{\prime \prime }} \sum^{T^{\prime \prime }}_{i=1} \rho_{i}\right)  \in \mathbb{R}^{T^{\prime \prime }} $ \\ 
        \quad\quad where $\rho =\max_{j\in [1,T^{\prime \prime }]} \mathcal{D}_{p}(i,j)$
        \STATE  {\# Compute the frame loss}
        \STATE $\mathcal{L}_{fr} \gets \mathcal{D}_{p}, \mathcal{D}_{n} $ by using \cref{triplet_loss}, (\ref{reg_loss}), and (\ref{frame_loss})
        \STATE \textbf{return} $Y_{sa}, \mathcal{L}_{fr}$ 
        \end{algorithmic}
    \end{algorithm}
        
    \subsection{Architecture Details}\label{arch}
        This section gives more details on the pipelines depicted in the main script to facilitate reproducibility and comprehension of the proposed system. These details describe the size of the data traveling through the pipeline in each module as well as the parameters for each layer. \Cref{fig:detail_pip_vvs} describes the detailed pipeline for the overall framework; \Cref{fig:detail_pip_ddm} for DDM; \Cref{fig:detail_pip_tsm} for TSM; and \Cref{fig:detail_pip_tgm} for TGM (on pages 13 and 14).

\section{Additional Ablation Studies \& Analyses}\label{ablation_analyses}
    \subsection{Quantitative Analysis on Ideal Suppression}\label{quan_exp}
        
\begin{figure}[!t]
        \centering
        \includegraphics[width=1\linewidth]{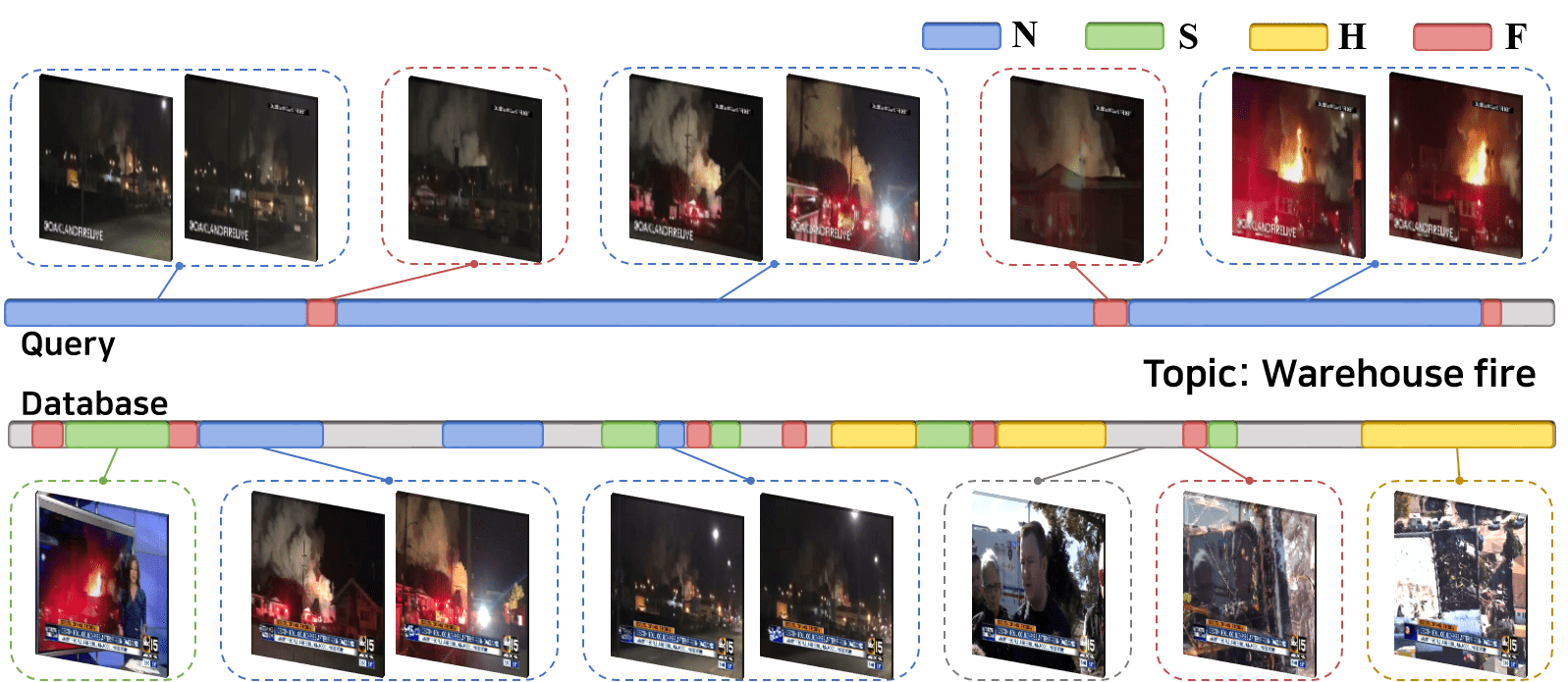} 
        \caption{\textbf{Example of Temporal Annotation for FIVR-5K.} This example illustrates the temporal annotations for a topic-related pair within FIVR-5K. N, S, H, and F represent segment-level labels for the related areas of the pair. N refers to segments where the temporal span and camera viewpoint are similar; S refers to segments where the temporal span is similar but the camera viewpoint is different; H refers to segments where the temporal span and camera viewpoint are different, but it can be inferred semantically as the same topic; and F refers to segments where a fading effect is observed before or after the segments corresponding to the preceding labels.\vspace{0mm}} \label{fig:temporal_ann}
    \end{figure}

        In this section, we demonstrate an ideal experiment to highlight the value of suppressing irrelevant frames when describing distinct video-level features. Concretely, the purpose of this experiment is to reveal how much previous approaches~\cite{kordopatis2017near, shao2021temporal, ng2022vrag} can profit from manually eliminating frames corresponding to distractors in their scheme. The experiment is performed on the FIVR-5K~\cite{kordopatis2019fivr} using temporal annotations to erase distractors. As depicted in~\Cref{fig:temporal_ann}, the temporal annotations~\cite{10077393} comprise segment-level labels indicating which areas of each video are related to one another for a topic-related pair between query and database in the FIVR-5K. The segment-level labels consist of three types: N, S, and H; the closer to N, the more visually related; and the closer to H, the more semantically related. In addition, this annotation also includes label F, indicating that the fade effect occurred before and after the three preceding labels. The remaining regions, excluding N, S, and~H, are presumed to be distractors because they contain irrelevant content to the topic or may cause confusion owing to the fade effect. As a consequence, for a total of 1,981 videos with the temporal annotations inside the FIVR-5K, video-level features are described while removing frames from locations identified as distractors in the scheme of earlier approaches. Note that 3,069 videos without the temporal annotations are handled in the original manner. Because this experiment employs the temporal annotations directly for prediction, it involves ``cheating'' and cannot be used as a benchmark. However, it provides a glimpse of the upper bound for existing approaches when optimal suppression is possible; it is dubbed Oracle, the same as in~\cite{huang2018makes}.

        \begin{figure}[!t]
            \centering
            \includegraphics[width=1\linewidth]{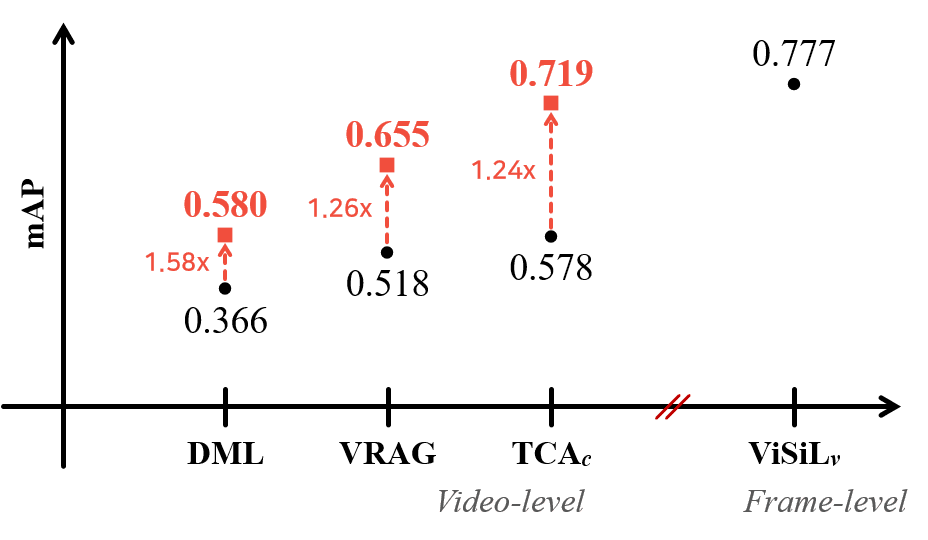} 
            \caption{\textbf{Oracle of Previous Approaches on FIVR-5K.} This indicates that proper suppression is key for the ideal video-level scheme. All performances are measured in the ISVR of FIVR-5K, with black circles representing the original performances of earlier approaches and red squares the performances after irrelevant frames are manually eliminated. \vspace{0mm}} \label{fig:oracle}
        \end{figure}

        \Cref{fig:oracle} shows the ISVR performance on FIVR-5K for the earlier video-level feature-based approaches obtained through the experiment stated above, i.e.~the Oracle. All cases when the temporal annotations are employed for ideal suppression are enhanced by $1.24$~times at least and $1.58$~times at most. In particular, the search accuracy, which has~been recognized as a weakness of video-level approaches, is~close to that of the most advanced frame-level approach~($\textup{ViSiL}_{v}$), implying that a scenario capable of fast and accurate response can be reached if only inappropriate frames are excluded when describing video-level features.

    \begin{figure*}[!t]    
        \centering
        \includegraphics[width=1\linewidth]{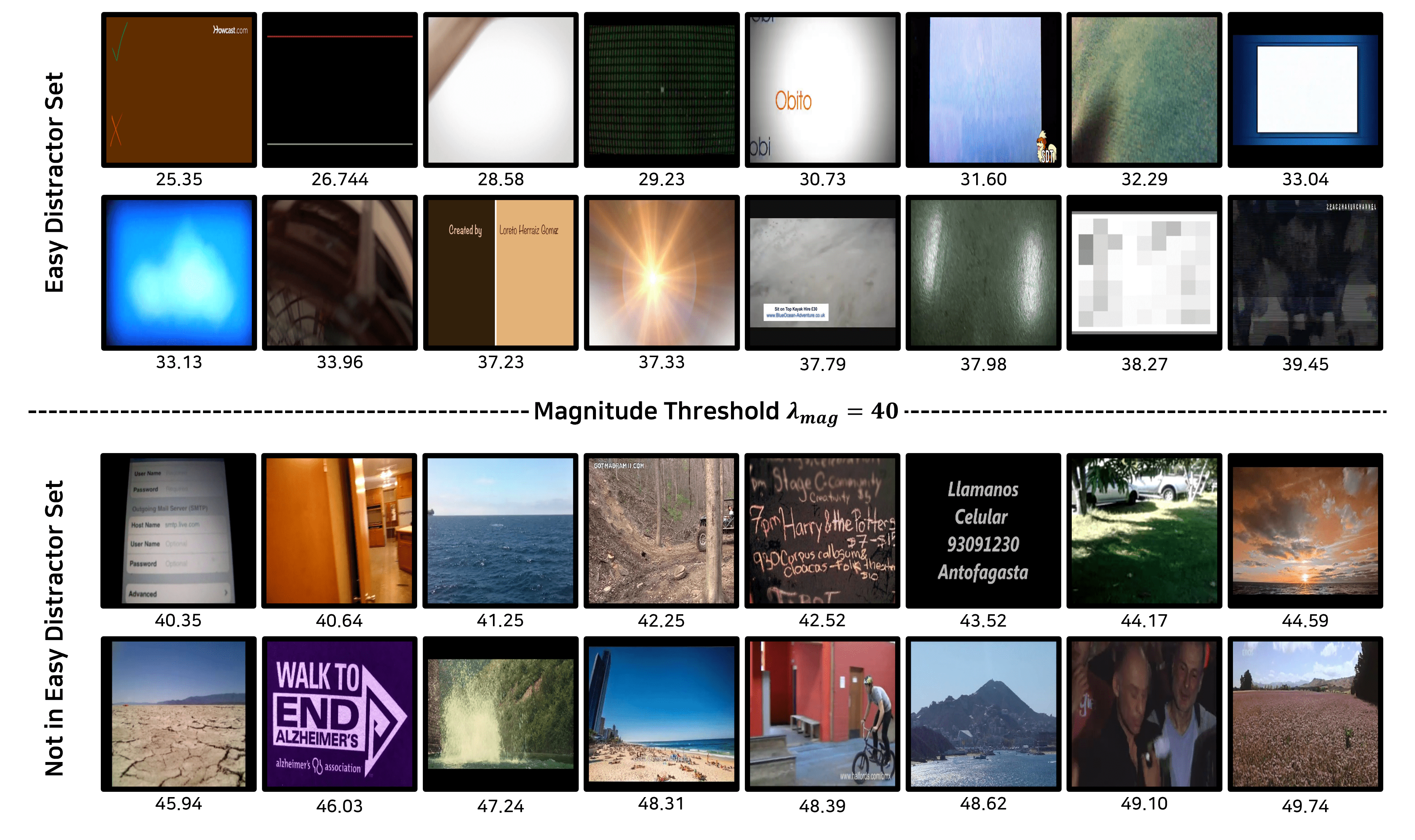} \vspace{0mm}
        \caption{\textbf{Example of Per-Frame Magnitude for Easy Distractor Selection on VCDB.} This represents the magnitude of frame-level features ($\textup{L}_{N}\textup{-iMAC}$) belonging to the metadata-free videos within the training dataset VCDB. The letters under each frame indicate its magnitude.\vspace{0mm}} \label{fig:ex_mag}
    \end{figure*}
    
    \begin{table}[!t]
        \footnotesize
        \centering
        \centering
        \begin{tabular}{c|c}
            \cmidrule[\heavyrulewidth]{1-2}
            \morecmidrules
            \cmidrule[\heavyrulewidth]{1-2}
            \begin{tabular}[c]{@{}c@{}}Magnitude\\Interval\end{tabular} & \begin{tabular}[c]{@{}c@{}}Cumulative \\ Percentage\end{tabular} \\ \midrule
               \:\:\;\;\:- 20          & \multicolumn{1}{l}{\;\;\;\:\:\:1.15\%}     \\ \midrule
                      20 - 30          & \multicolumn{1}{l}{\;\;\;\:\:\:2.61\%}     \\ \midrule
                      30 - 40          & \multicolumn{1}{l}{\;\;\;\:\:\:8.96\%}     \\ \midrule
                      40 - 50          & \multicolumn{1}{l}{\;\;\:\:25.19\%}        \\ \midrule
                      50 - 60          & \multicolumn{1}{l}{\;\;\:\:61.46\%}\       \\ \midrule
            \multicolumn{1}{l|}{\ \ \;60 -}         & 100.00\%                      \\
            \cmidrule[\heavyrulewidth]{1-2}
            \morecmidrules
            \cmidrule[\heavyrulewidth]{1-2}
        \end{tabular}
        \vspace{0mm}
        \caption{\textbf{Percentage by Magnitude Interval on VCDB.} This is the cumulative percentage by interval of the magnitude of all frame-level features derived from metadata-free videos within the VCDB. ``- 20" refers to an interval whose magnitude is less than 20, and ``60 -" refers to 60 or higher.} \label{tab:mag_cur}
    \end{table}

    \subsection{Examples within Easy Distractor Set} \label{ddm_exam}

        This section discusses the easy distractor set used for DDM training. The easy distractor set is derived from metadata-free videos in the training dataset VCDB~\cite{jiang2014vcdb}, as stated in the main script. Specifically, $\textup{L}_{N}\textup{-iMAC}$, which are frame-level features, are extracted from all of these videos, but PCA are not applied in order to assess directly to what extent layers of the backbone network activate low-level characteristics. After that, their magnitude is filtered by the magnitude threshold~$\lambda_{mag}$; if this value is less than the threshold, the frame is added to the easy distractor set. The reason for identifying easy distractors in this criterion is that fewer low-level characteristics and smaller pixel variations result in fewer elements being activated in the process of describing frame-level features, leading to a smaller scale of output elements. However, if filtering is performed with a magnitude threshold that is too small for a perfect easy distractor, the variance within the set of easy distractors is reduced, leading to overfitting of the pattern of a particular easy distractor. In contrast, if filtering is conducted with a value that is too high, many frames will be deemed to be the easy distractor and eliminated by DDM, despite the fact that they may include several signals for topic comprehension. To select an appropriate magnitude threshold value in this trade-off relationship, the magnitudes of all frame-level features are categorized by interval and converted to cumulative percentages, as shown in~\Cref{tab:mag_cur}. As a result, the magnitude threshold is set to 40 (corresponding to ``30-40"), considering that the cumulative amount of the interval is a sufficient proportion close to 10\% of the total training data before the cumulative percentage exploded rapidly. In addition, the appropriateness of this value is illustrated with qualitative examples in~\Cref{fig:ex_mag}; in fact, when the magnitude threshold is less than 40, it is evident that the majority of frames are assessed as easy distractors. Even though there are frames that can be identified as easy distractors when the magnitude threshold is more than 40, there are also numerous landscapes that may be used to depict the topic of a certain video, demonstrating that the magnitude threshold currently in use is a well-balanced value.

    \begin{table}[!t]
    \footnotesize
        \small
        \centering
        \begin{tabular}{cccccc}
            \cmidrule[\heavyrulewidth]{1-6}
            \morecmidrules
            \cmidrule[\heavyrulewidth]{1-6} 
             & \multirow{2}{*}[-.25em]{\begin{tabular}[c]{@{}c@{}}Manual\\ \textit{Thr.}\end{tabular}} & \multirow{2}{*}[-.25em]{\textit{Mag.}} & \multicolumn{3}{c}{FIVR-5K} \\\cmidrule(){4-6}
             &  &  & DSVR & CSVR & ISVR \\ \midrule
            Baseline &  &  & 0.692 & 0.700 & 0.651 \\ \midrule
            DDM only &  & 40 & \textbf{0.715} & \textbf{0.725} & \textbf{0.672} \\ \midrule
            \multirow{7}{*}{\begin{tabular}[c]{@{}c@{}}\textit{w/o}\\ Training\end{tabular}} & \cmark & 10 & 0.692 & 0.700 & 0.651 \\
             & \cmark & 20 & 0.692 & 0.700 & 0.651  \\
             & \cmark & 30 & 0.696 & 0.705 & 0.656  \\
             & \cmark & 40 & 0.703 & 0.712 & 0.663  \\
             & \cmark & 50 & 0.706 & 0.715 & 0.665  \\
             & \cmark & 60 & 0.672 & 0.684 & 0.633  \\
             & \cmark & 70 & 0.468 & 0.480 & 0.454  \\
            \cmidrule[\heavyrulewidth]{1-6}
            \morecmidrules
            \cmidrule[\heavyrulewidth]{1-6}              
        \end{tabular}
    \caption{\textbf{Benefits of Training DDM for Suppression.} Manual \textit{Thr.} represents the use of manual elimination with thresholding via the magnitude of frame-level features during the inference phase without any training. \textit{Mag.} denotes the magnitude threshold; in DDM only, it is used in the training phase, as mentioned in the main script. Baseline represents the method that weighs all frames equally without any of the proposed modules (referred to in the module-wise ablation studies of $\textup{VVS}_{\small{\textit{3840}}}$ presented in the main script).
\vspace{-0mm}}
    \label{tab:trainingddm}
    \end{table}
    
    \subsection{Benefits of Training DDM for Suppression} \label{training_ddm}
        This section explains the reason for training DDM as opposed to simply observing the magnitude of frame-level features to remove easy distractors in the inference phase. First, it is important to note that directly adjusting the magnitude threshold during the inference phase for optimal performance on each dataset can be considered ``cheating". This is because in the actual scenario where the retrieval service is provided, this adjustment is not possible because the correct answer to the evaluation data is not known in advance. Furthermore, as indicated in~\Cref{tab:trainingddm}, training DDM for suppression demonstrates a higher improvement compared to manual removal without any training. The results suggest that optimizing the recognition of easy distractors using magnitude thresholding during the training phase, like in DDM, can be performed robustly even if there are some noises in the thresholding; on the other hand, adjusting the magnitude threshold directly during the inference phase is more vulnerable to such noise.

    \begin{table}[t]
        \footnotesize
        \centering
        \begin{tabular}{ccc|ccc}
        \cmidrule[\heavyrulewidth]{1-6}
        \morecmidrules
        \cmidrule[\heavyrulewidth]{1-6}
        \multicolumn{3}{c|}{Tempered Sigmoid} & \multicolumn{3}{c}{FIVR-5K}                      \\ \midrule
        DDM        & TSM        & TGM         & DSVR           & CSVR           & ISVR           \\ \midrule
                   &            &             & 0.727          & 0.744          & 0.701          \\ \midrule
        \cmark     &            &             & 0.728          & 0.739          & 0.694          \\
                   & \cmark     &             & 0.737          & 0.743          & 0.696          \\
        \textbf{}  & \textbf{}  & \textbf{\cmark}  & \textbf{0.744} & \textbf{0.752} & \textbf{0.705} \\ \midrule
        \cmark     & \cmark     & \cmark      & 0.733          & 0.740          & 0.695          \\
        \cmidrule[\heavyrulewidth]{1-6}
        \morecmidrules
        \cmidrule[\heavyrulewidth]{1-6}
        \end{tabular}
        \vspace{0mm}
        \caption{\textbf{Effect of Tempered Sigmoid on Each Module.} If a check mark is present, it indicates that the tempered sigmoid replaces the sigmoid of that module.} \label{tab:tempered}
    \end{table}
    
    \subsection{Advantages of Tempered Sigmoid} \label{tempered_exp}
        This section discusses why the tempered sigmoid is used instead of the sigmoid only in TGM. The tempered sigmoid~\cite{papernot2021tempered} was proposed in order to provide robustness to noise in the training phase by controlling the gradient norm, which can be formulated as follows:
        \begin{equation}
            \small
            \omega(h) = \frac{\sigma }{1+e^{-h/\tau}}-o
            \label{eq:tempered}
        \end{equation}
        where $\omega$ refers to the tempered sigmoid function, $\sigma$ to a parameter that controls the activation scale, $\tau$ to the temperature that reduces the gradient norm so that model parameters can be updated carefully, and $o$ to the offset.

        In fact, TGM receives as input the output of DDM, which comprises injected easy distractors as well as topic-related and topic-unrelated content that originally existed in an input video. As the ST-GAP, a global operation, is applied to the input of TGM to obtain the rough pattern, noise has a greater impact on optimization than other modules. Therefore, the tempered sigmoid with the aforementioned benefits is utilized to train TGM more stably. Here, $\sigma$ and $o$ are set to $1$ and $0$ because the result of the tempered sigmoid is utilized guidance weights, which have values between $0$ and $1$. Also, $\tau$ is empirically set to~512.
        
        As seen in~\Cref{tab:tempered}, when the tempered sigmoid is used instead of the sigmoid solely for TGM, the performance of the ISVR increases from 0.701 to 0.705 compared to when the tempered sigmoid is not employed anywhere. However, as other modules besides TGM have a relatively low emerging chance of noise, a small gradient norm caused by the tempered sigmoid only results in insufficient optimization.

    \subsection{Effect of Saliency Signal on Representation} \label{effect_saliency_signal}
       
        In TSM, the model is guided by a pseudo-label, i.e., a saliency label, generated based on saliency signals (also referred to as saliency information) within a frame-level similarity map. This section covers a direct interpretation of whether these saliency signals actually contribute to the representation of video-level features.
    
\begin{figure}[!t]
            \begin{subfigure}{\linewidth}
                \centering
                \includegraphics[width=1\linewidth]{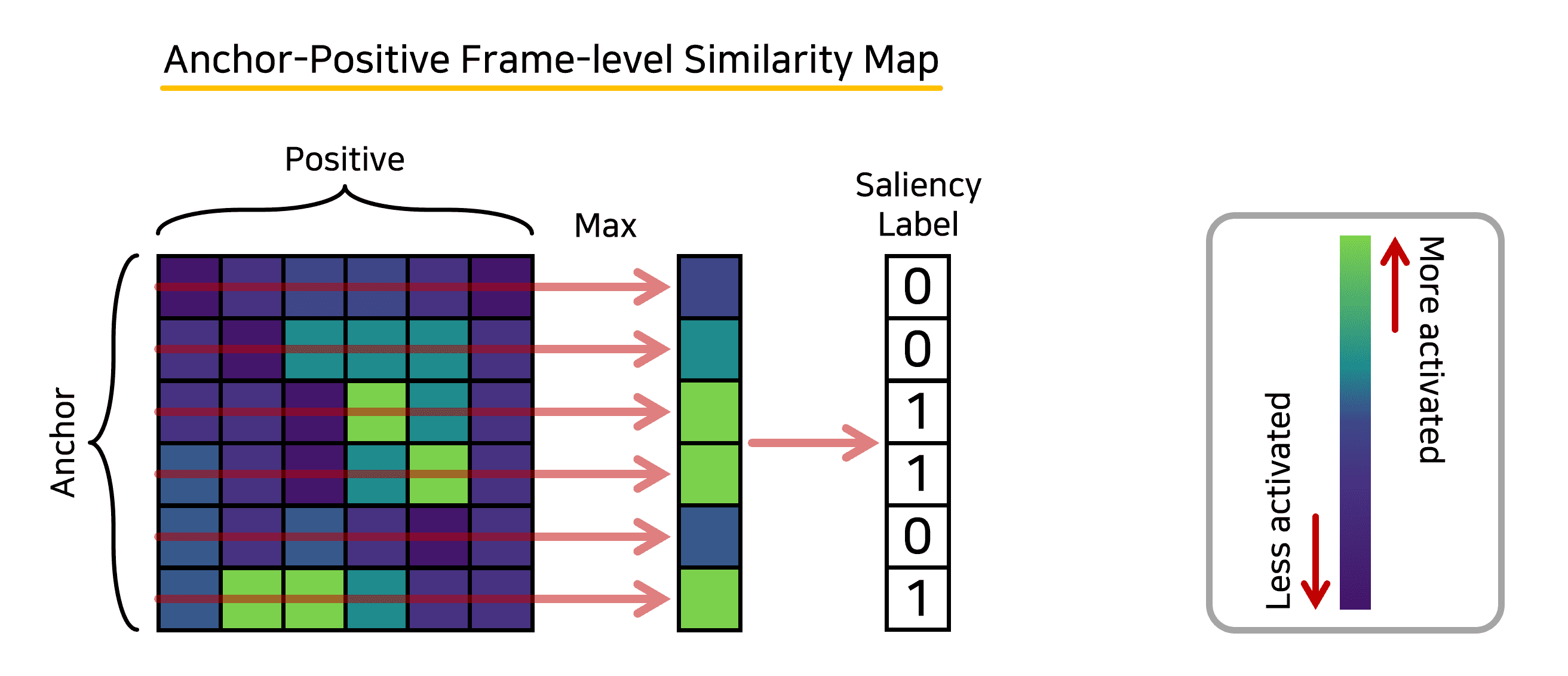}
                \caption{Process for Calculating Saliency Label}
                \label{fig:dw_sub1}
            \end{subfigure}
            \begin{subfigure}{\linewidth}
            \vspace{6mm}
                \centering
                \includegraphics[width=1\linewidth]{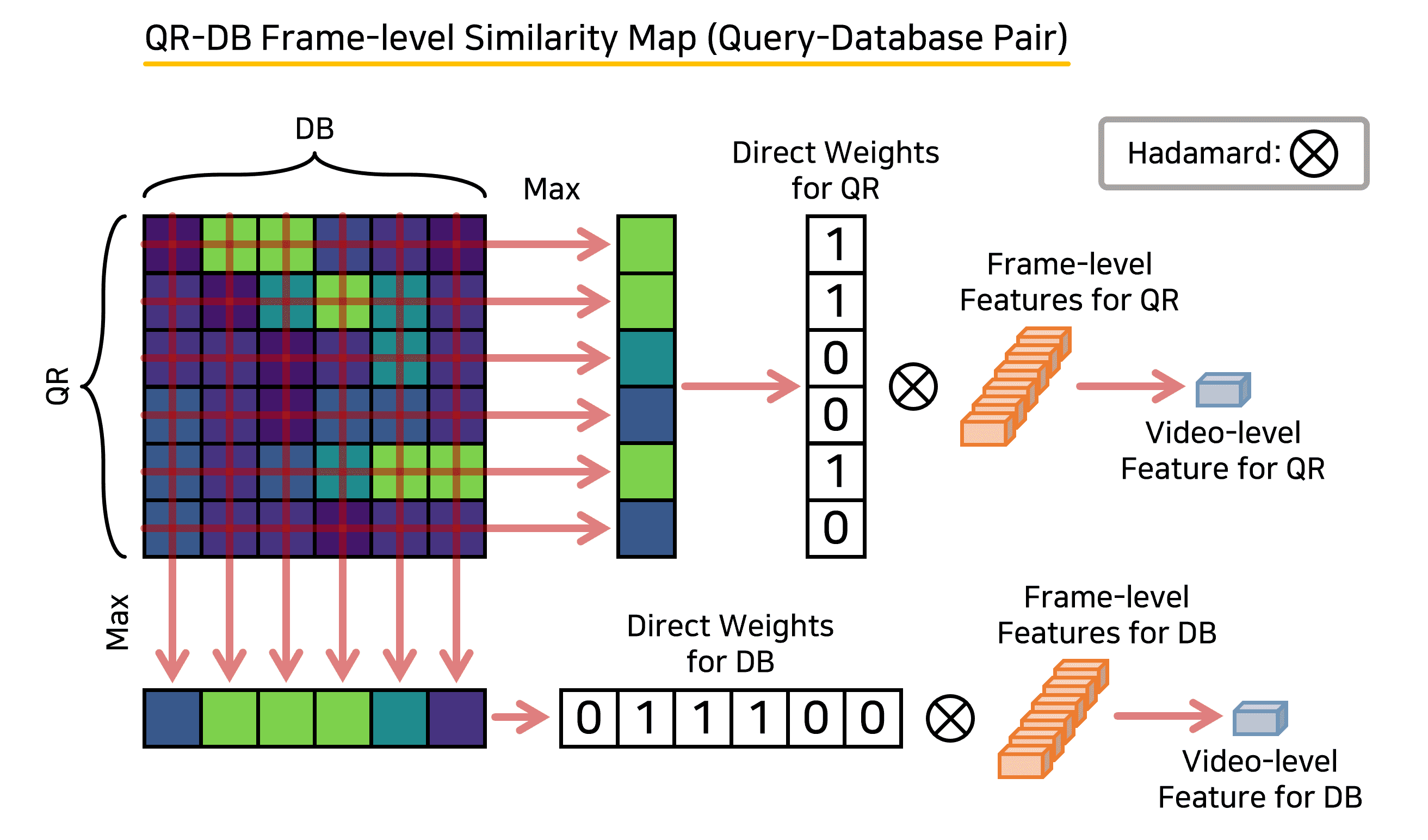}
                \caption{Process for Calculating Direct Weights}
                \label{fig:dw_sub2}
            \end{subfigure}
    
            \caption{\textbf{Toy Example for Direct Weights.} This simplifies the calculation process of direct weights for demonstrating the effect of saliency signals on video-level feature representation, as discussed in~\Cref{effect_saliency_signal}, in comparison to the calculation process of saliency label.}\vspace{0mm}
            \label{fig:dw_all}
        \end{figure}

        To verify the contribution of saliency signals to the representation, direct weights obtained from a frame-level similarity map are employed during the inference phase. The process begins with the baseline (referred to in the module-wise ablation studies of $\textup{VVS}_{\small{\textit{3840}}}$ presented in the main script), in which all frame-level features are aggregated with the same weights without any of the proposed modules. Technically, before describing a video-level feature, as shown in (b) of~\Cref{fig:dw_all}, a frame-level similarity map is computed for each pair consisting of one query and one video within the database. Here, the map is identical to when acquiring a saliency label, i.e., it is a finely tuned map. The maximum operation is then applied to the map and converted to 1 or 0, which is also the same as obtaining a saliency label (this process is illustrated in (a) of~\Cref{fig:dw_all}), but the outcome here corresponds to direct weights for the query. In addition, similar to this procedure, the maximum operation is applied in an orthogonal direction to produce direct weights for the video within the database. After that, a video-level feature for each video is aggregated by using the Hadamard product between the frame-level features of each video and the direct weights for that video.

    \begin{table}[!t] 
        \footnotesize
        \centering
        \begin{tabular}{c|ccc}
            \cmidrule[\heavyrulewidth]{1-4}
            \morecmidrules
            \cmidrule[\heavyrulewidth]{1-4}  
            \multirow{2}{*}[-.3em]{\begin{tabular}[c]{@{}c@{}}\end{tabular}} & \multicolumn{3}{c}{FIVR-5K} \\ \cmidrule(){2-4}
               & DSVR  & CSVR  & ISVR  \\ \cmidrule(){1-4} 
               (a) & 0.692 & 0.700 & 0.651 \\
              (b) & \textbf{0.799} & \textbf{0.799} & \textbf{0.726} \\
             \cmidrule[\heavyrulewidth]{1-4}
            \morecmidrules
            \cmidrule[\heavyrulewidth]{1-4}
        \end{tabular}
        \vspace{0mm}
        \caption{\textbf{Effect of Saliency Signal on Representation.} (a) represents the baseline (referred to in the module-wise ablation studies of $\textup{VVS}_{\small{\textit{3840}}}$ presented in the main script) that weighs all frames equally without any of the proposed modules. (b) represents the scenario when direct weights are employed at the baseline, as stated in~\Cref{effect_saliency_signal}. \vspace{0mm}}
        \label{tab:effect_saliency_signal}
    \end{table}

        This verification procedure cannot be termed a ``video-level approach" because the similarities are computed first from frame-level features in order to create the direct weights during the inference phase. However, this procedure allows one to figure out to what extent saliency signals can help describe video-level features. Because the direct weights for the query and the video within the database, which are formed in the same manner as a saliency label, are directly engaged in aggregating video-level features.

        As shown in~\Cref{tab:effect_saliency_signal}, when direct weights are employed~(b), significant performance increases are found over the baseline~(a), indicating that saliency signals are strongly linked to strengthening the representation of video-level features. Moreover, the results in that table demonstrate that the saliency label in TSM mainly leads the model to explore visual correlations, as the improvements in the DSVR task, where discovering visually related scenes is a priority, are greater than in other tasks.

    \subsection{Effect of Hierarchical Connection on TGM}\label{hcon_eff} 
    
        This section explains the effectiveness of leveraging the hierarchical connection in TGM. To recap what was discussed in the main script, the hierarchical connection is a strategy to assist the model in grasping the topic by directly utilizing data that covers different temporal spans caused by multiple convolution layers in the refinement process. This direct use of knowledge on various temporal spans is intended to help understand the content of varying lengths included in videos, and it can be expected that this strategy will be more beneficial for longer videos, which have more content.

    \begin{figure}[t!]
        \begin{subfigure}{\linewidth}
            \centering
            \includegraphics[width=1\linewidth]{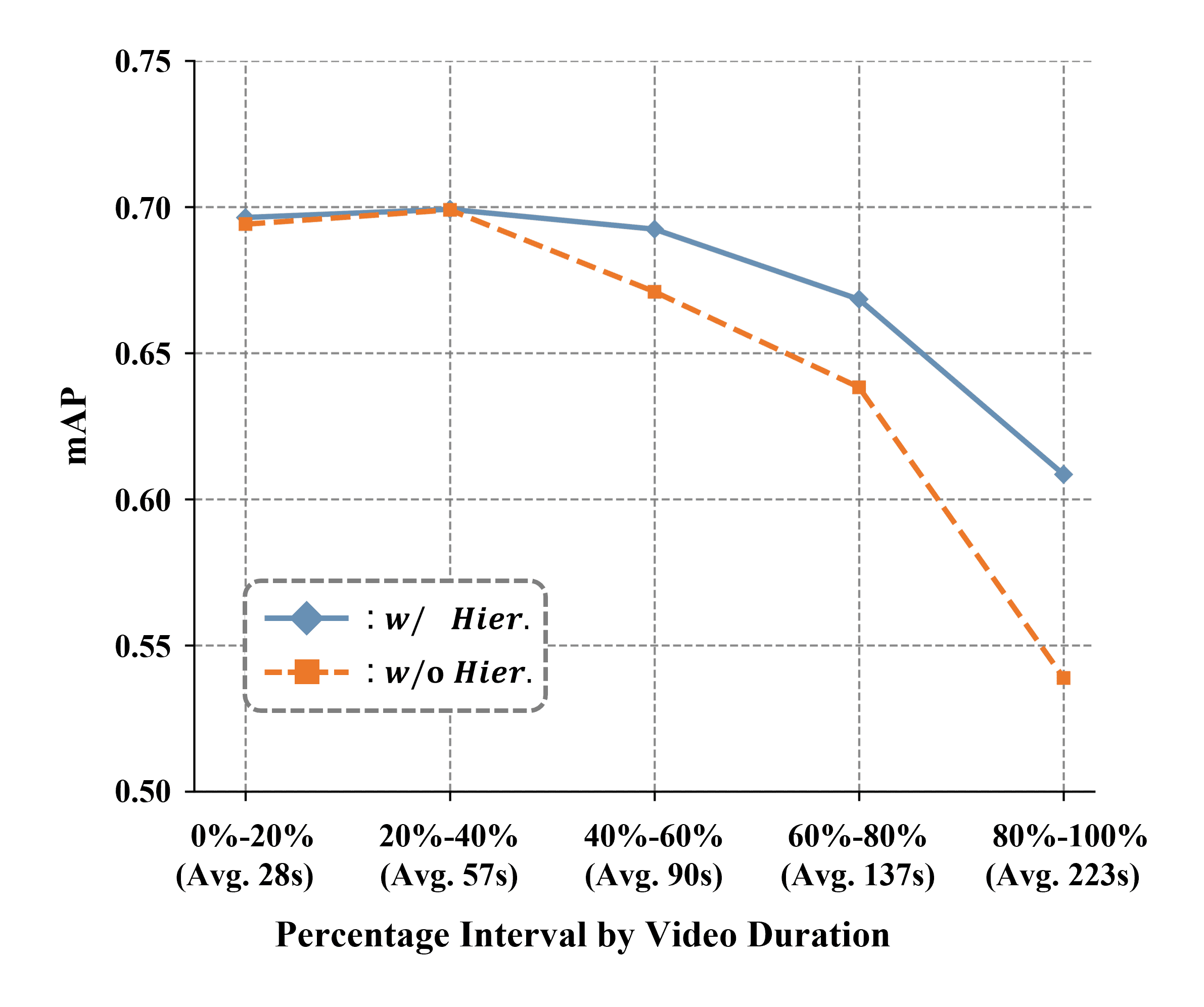}
            \caption{Performance over Duration on FIVR-5K}
            \label{fig:hc_sub1}
        \end{subfigure}
        \begin{subfigure}{\linewidth}
        \vspace{6mm}
            \centering
            \includegraphics[width=1\linewidth]{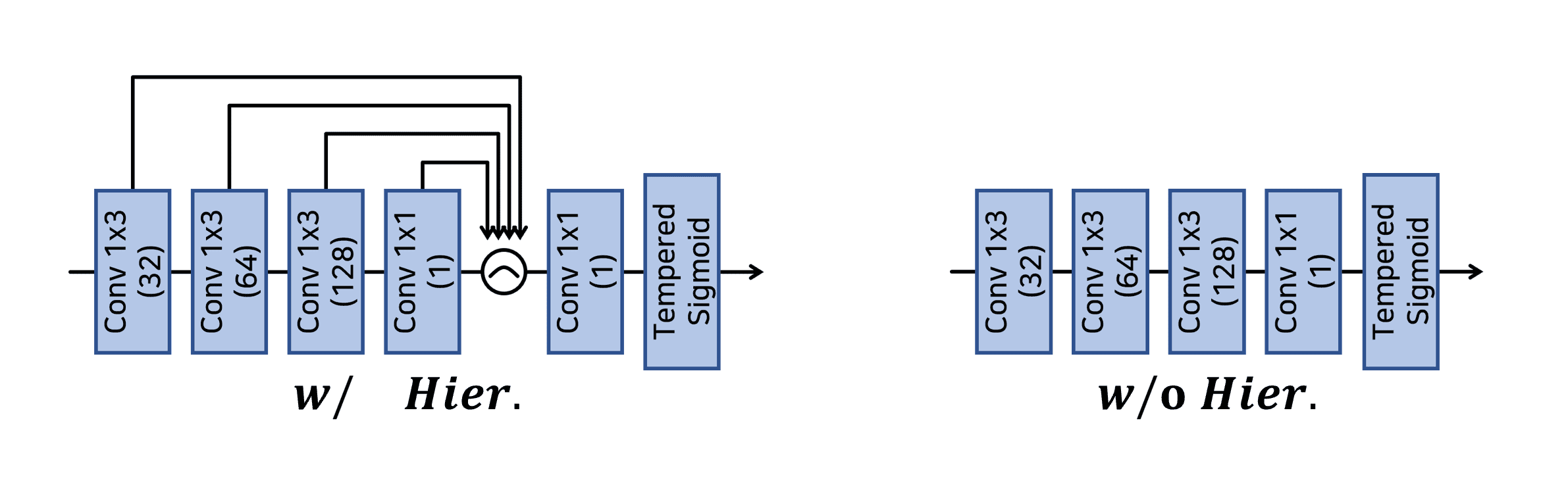}
            \caption{Structural Difference of TGM by Hierarchical Connection}
            \label{fig:hc_sub2}
        \end{subfigure}
        \caption{\textbf{Effect of Hierarchical Connection.} This demonstrates the merit of hierarchical connection within the TGM according to various video durations on FIVR-5K. In~(a), the horizontal axis indicates that each of the five subsets into which the FIVR-5K database is divided based on the duration interval replaces the entire database. In other words, 0\%-20\% is a case where only one-fifth of the database is used in the order of shorter duration in the database, and 80\%-100\% is a case in the order of longer duration. Furthermore, the average duration of the videos included in each interval is indicated in parentheses below each one. The vertical axis represents mAP in ISVR, and \textit{Hier.} refers to the hierarchical connection. A blue solid line indicates when the hierarchical connection exists, while an orange dashed line indicates when it does not. The structural difference in TGM resulting from the presence of hierarchical connection can be seen in~(b).\vspace{0mm}}
        \label{fig:hc_all}
    \end{figure}

        In fact, the experiment represented in (a) of~\Cref{fig:hc_all} demonstrates that the expectation is satisfied. The experiment is conducted by splitting the full database such that it contains just videos of a particular duration interval and evaluating the effect of hierarchical connection on that duration interval. Specifically, the FIVR-5K database is separated into five subsets of the same amount based on the duration order. The search performance is then evaluated using each subset as a database. At this time, the performances over the duration interval are measured for the presence or absence of the hierarchical connection. Furthermore, their structural difference can be seen in (b) of~\Cref{fig:hc_all}. Consequently, the longer videos, the higher the performances when the hierarchical connection is present as opposed to when it is not. This suggests that, in accordance with the design intent of the hierarchical connection, the direct integration over multiple temporal spans in lengthy videos with several contents of varying durations aids comprehension of the topic.

    \subsection{Efficiency on Speed and Memory} \label{speed_memory}
        \Cref{tab:speed_mem} demonstrates that the video-level feature-based approach is much faster than the frame-level approach in terms of average inference seconds per query and that, regardless of dimensions, our approach is at least 3.3~times faster than the fastest frame-level approach ($\textup{TCA}_{f}$) on the FIVR-5K. Furthermore, video-level approaches like ours store about 108 times (this ratio increases as average duration in the dataset increases) less on this dataset. This is quite memory efficient considering the actual scenario in which the database features are stored in advance in memory to facilitate the processing of queries transmitted in real time. In summary, this section demonstrates that our approach, which is close to the accuracy of frame-level approaches as presented in the main script, is also competitive in terms of speed and memory efficiency, which are the primary characteristics of video-level approaches.

    \begin{table}[!t] 
        \begin{center}
        \footnotesize
        \begin{tabular}{@{}clrrc@{}}
            \cmidrule[\heavyrulewidth]{1-5}
            \morecmidrules
            \cmidrule[\heavyrulewidth]{1-5} 
            \multirow{2}{*}{} & \multicolumn{1}{c}{\multirow{2}{*}[-.0em]{Approach}} & \multicolumn{1}{c}{\multirow{2}{*}[-.0em]{\textit{Dim.}}} &
            \multirow{2}{*}[-.0em]{\begin{tabular}[c]{@{}c@{}}Inference Time \\ (sec/q)\end{tabular}} & \multirow{2}{*}[-.0em]{\begin{tabular}[c]{@{}c@{}} \# of \\ Features \end{tabular}}\\ 
            &  &  &  &  \\ \midrule
            \multicolumn{1}{c}{\multirow{7}{*}{\rotatebox[origin=c]{90}{\textit{\textbf{frame}}}}}
             & DP & \multicolumn{1}{c}{\;\;-}  & 137.57s\;\;\;\;\, & \multicolumn{1}{c}{\multirow{7}{*}{540,361}} \\
            \multicolumn{1}{l}{} & TN & \multicolumn{1}{c}{\;\;-} & 3.61s\;\;\;\;\, &  \\
            \multicolumn{1}{l}{} & $\textup{ViSiL}_{sym}$ & 3,840 & 8.24s\;\;\;\;\, &  \\
            \multicolumn{1}{l}{} & $\textup{ViSiL}_{v}$  & 3,840 & 8.16s\;\;\;\;\, &  \\
            \multicolumn{1}{l}{} & $\textup{ViSiL}_{f}$  & 3,840 & 3.76s\;\;\;\;\, &  \\
            \multicolumn{1}{l}{} & $\textup{TCA}_{sym}$  & 1,024 & 4.00s\;\;\;\;\, &  \\
            \multicolumn{1}{l}{} & $\textup{TCA}_{f}$  & 1,024 & 2.14s\;\;\;\;\, &  \\ 

             \midrule 
            \multicolumn{1}{c}{\multirow{8}{*}[0em]{\rotatebox[origin=c]{90}{\textit{\textbf{video}}}}} 
            & TMK  & 65,536 & 7.23s\;\;\;\;\, & \multicolumn{1}{c}{\multirow{8}{*}{\textbf{5,000}}} \\ 
            & VRAG  & 4,096 & 0.79s\;\;\;\;\, &  \\
             & DML  & 500 &  0.61s\;\;\;\;\, &  \\ 
             
             & $\textup{TCA}_{c}$  & 1,024 & 0.28s\;\;\;\;\, & \\ \cmidrule(lr){2-4}
             & $\textbf{\textup{VVS}}_{\textit{500}}$ \,\,\textbf{(ours)} & 500 & \textbf{0.57s}\;\;\;\;\, &  \\
             & $\textbf{\textup{VVS}}_{\textit{512}}$ \,\,\textbf{(ours)} & 512 & \textbf{0.58s}\;\;\;\;\, & \\
             & $\textbf{\textup{VVS}}_{\textit{1024}}$ \textbf{(ours)} & 1,024 & \textbf{0.59s}\;\;\;\;\, & \\
             & $\textbf{\textup{VVS}}_{\textit{3840}}$ \textbf{(ours)} & 3,840 & \textbf{0.64s}\;\;\;\;\, &  \\
            \cmidrule[\heavyrulewidth]{1-5}
            \morecmidrules
            \cmidrule[\heavyrulewidth]{1-5} 
            \end{tabular}
            \vspace{0mm}
            \caption{\textbf{Inference Time \& Number of Features.} This demonstrates the average inference seconds per query on the FIVR-5K and the number of those database features that need to be stored in memory. The \textit{\textbf{frame}} and \textit{\textbf{video}} refer to frame-level and video-level feature-based approaches. \textit{Dim.}~refers to the dimension of the basic unit for calculating similarity in each approach (i.e., frame-level approaches use multiple features of that dimension, as many as the number of all or most frames in a video, while video-level approaches use only one feature of that dimension). } \vspace{0mm} \label{tab:speed_mem}
        \end{center}
    \end{table}

    \begin{table*}[!t] 
        \begin{center}
        \footnotesize
        \setlength{\tabcolsep}{4pt}
        \begin{tabular}{@{}clrcccccccccccccc@{}}
            \cmidrule[\heavyrulewidth]{1-17}
            \morecmidrules
            \cmidrule[\heavyrulewidth]{1-17} 
            \multirow{2}{*}{} & \multicolumn{1}{c}{\multirow{2}{*}[-.3em]{Approach}} & \multicolumn{1}{c}{\multirow{2}{*}[-.3em]{\textit{Dim.}}} &
            \multicolumn{14}{c}{EVVE} \\ \cmidrule(l){4-17} 
            &  & & \textit{Avg.} & \#1 & \#2 & \#3 & \#4 & \#5 & \#6 & \#7 & \#8 & \#9 & \#10 & \#11 & \#12 & \#13 \\ \midrule
            
            \multicolumn{1}{c}{\multirow{5}{*}[0em]{\rotatebox[origin=c]{90}{\textit{\textbf{frame}}}}} & Jo \textit{et al.} & 1,040 & 0.559 & 0.749 & 0.602 & 0.315 & 0.318 & 0.355 & 0.316 & 0.336 & 0.229 & 0.302 & 0.462 & 0.176 & 0.843 & 0.645 \\ 
            \multicolumn{1}{l}{} & $\textup{TCA}_{f}$ & 1,024 & 0.636 & 0.885 & 0.698 & 0.251 & \textbf{0.449} & 0.395 & \textbf{0.390} & 0.289 &	0.271 &	0.583 &	0.666 &	0.338 &	0.893 &	\textbf{0.829} \\
            \multicolumn{1}{l}{} & $\textup{ViSiL}_{f}$ & 3,840 & 0.585 &	0.834 &	0.625 &	0.148 &	0.427 &	0.347 &	0.355 &	0.277 &	0.183 &	0.355 &	0.586 &	0.296 &	0.860 &	0.747 \\   
            \multicolumn{1}{l}{} & $\textup{ViSiL}_{sym}$ & 3,840 & 0.646 &	0.858 &	0.775 &	\textbf{0.434} &	0.399 &	\textbf{0.418} & 0.298 &	\textbf{0.371} &	\textbf{0.295} &	\textbf{0.698} &	0.583 &	0.315 &	\textbf{0.930} &	0.762 \\
            \multicolumn{1}{l}{} & $\textup{ViSiL}_{v}$ & 3,840 & \textbf{0.659} &	\textbf{0.919} &	\textbf{0.810} &	0.360 &	0.421 &	0.405 &	0.360 &	0.298 &	0.278 &	0.567 &	\textbf{0.667} &	\textbf{0.391} &	0.929 &	0.827 \\ \midrule 
            \multicolumn{1}{c}{\multirow{7}{*}[0em]{\rotatebox[origin=c]{90}{\textit{\textbf{video}}}}} & DML & 500 & 0.541 &	0.414 &	0.461 &	0.082 &	0.233 &	0.315 &	0.300 &	0.259 &	0.170 &	0.078 &	0.470 &	0.274 &	0.883 &	0.705 \\ 
             & $\textup{TCA}_{c}$ & 1,024 & 0.599 &	0.687 &	0.591 &	0.154 &	0.390 &	0.357 &	0.381 &	0.288 &	0.264 &	0.544 &	0.615 &	0.272 &	0.871 &	0.790 \\
             & VRAG & 4,096 & 0.632 &	0.772 &	0.705 &	0.104 &	0.283 &	0.370 &	0.311 &	0.286 &	\textbf{0.302} &	0.610 &	\textbf{0.701} &	0.371 &	\textbf{0.918} &	0.762 \\\cmidrule(l){2-17}
             & $\textbf{\textup{VVS}}_{\textit{500}}$ \,\,\textbf{(ours)} & 500 & 0.629 &	0.732 &	0.659 &	\textbf{0.210} &	0.398 &	0.351 &	0.409 &	0.292 &	0.274 &	0.588 &	0.684 &	0.343 &	0.904 &	0.811 \\
             & $\textbf{\textup{VVS}}_{\textit{512}}$ \,\,\textbf{(ours)} & 512 & 0.630 &	0.733 &	0.654 &	0.206 &	0.413 &	0.353 &	0.416 &	0.289 &	0.273 &	\textbf{0.611} &	0.689 &	0.348 &	0.903 &	0.812 \\
             & $\textbf{\textup{VVS}}_{\textit{1024}}$ \textbf{(ours)} & 1,024 & 0.630 &	0.763 &	0.669 &	0.168 &	0.396 &	0.360 &	0.401 &	0.304 &	0.285 &	0.516 &	0.683 &	0.353 &	0.902 &	0.807 \\
             & $\textbf{\textup{VVS}}_{\textit{3840}}$ \textbf{(ours)} & 3,840 & \textbf{0.644} &	\textbf{0.835} &	\textbf{0.731} &	\textbf{0.210} &	\textbf{0.433} &	\textbf{0.375} &	\textbf{0.423} &	\textbf{0.325} &	0.282 &	0.423 &	0.667 &	\textbf{0.403} &	0.910 &	\textbf{0.840} \\
            \cmidrule[\heavyrulewidth]{1-17}
            \morecmidrules
            \cmidrule[\heavyrulewidth]{1-17} 
            \end{tabular}%
            \vspace{0mm}
        \caption{\textbf{Benchmark on EVVE.} The results are reported on a subset we own ($\approx$70.5\% of the original) using the trained model provided by the official code of each approach due to the unavailability of the full original dataset. The \textit{\textbf{frame}} and \textit{\textbf{video}} refer to frame-level and video-level feature-based approaches. \textit{Dim.}~refers to the dimension of the basic unit for calculating similarity in each approach (i.e., frame-level approaches use multiple features of that dimension, while video-level approaches use only one feature of that dimension). The subscript \textit{2} indicates that the feature of that dimension is binarized. \#1 through \#13 refer to the mAP of the event corresponding to the id listed in~\cite{revaud2013event}, and \textit{Avg.} refers to the mAP for all events. Only approaches that are trained from VCDB or do not require additional training are shown for a fair comparison.} \vspace{0mm} \label{tab:evve}
        \end{center}
    \end{table*}
    
    \subsection{Comparison with Others on EVVE} \label{bench_evve}

        \Cref{tab:evve} shows comparisons with earlier approaches on the EVVE~\cite{revaud2013event}, a dataset for event video retrieval (EVR). Note that several videos in the original dataset were missing; however, the benchmark is presented on the subset we own~($\approx$70.5\% of the original) for further comparison on a dataset considered more challenging.

        Despite having fewer dimensions, $\textup{VVS}_{\small{\textit{3840}}}$ outperforms VRAG~\cite{ng2022vrag}, which represents the state-of-the-art performance among current video-level approaches. This performance is better than the majority of the frame-level state-of-the-art. Furthermore, $\textup{VVS}_{\small{\textit{500}}}$, $\textup{VVS}_{\small{\textit{512}}}$ and $\textup{VVS}_{\small{\textit{1024}}}$ with dimension reduction during the PCA procedure surpass the video-level approach of the same dimension in every case.

    \begin{table*}[ht!]
        \begin{center}
        \footnotesize
        \begin{tabular}{llcccccl}
            \cmidrule[\heavyrulewidth]{1-8}
            \morecmidrules
            \cmidrule[\heavyrulewidth]{1-8} 
            
            \multicolumn{1}{c}{\multirow{2}{*}{Approach}} & \multicolumn{1}{c}{\multirow{2}{*}{Reference}} & \multicolumn{2}{c}{SumMe} & \multicolumn{2}{c}{TVSum} & \multicolumn{1}{c}{\multirow{2}{*}{\begin{tabular}[c]{@{}c@{}}Average \\ \textit{Rank.}\end{tabular}}} & \multicolumn{1}{c}{\multirow{2}{*}{\begin{tabular}[c]{@{}c@{}}Data\\ splits\end{tabular}}} \\ \cmidrule(lr){3-4} \cmidrule(lr){5-6} 
             &  & F-score & \textit{Rank.} & F-score & \textit{Rank.} & \\ \midrule
             Random summary             & \multicolumn{1}{c}{-} & 40.2 & 19 & 54.4 & 16 & 17.5 & \multicolumn{1}{c}{-} \\ \midrule
             $\textup{SUM-FCN}_{unsup}$ & \cite{rochan2018video}  & 41.5 & 17 & 52.7 & 17 & 17 & M Rand \\
             DR-DSN & \cite{zhou2018deep} & 41.4 & 18 & 57.6 & 13 & 15.5 & 5 Rand \\
             EDSN &  \cite{gonuguntla2019enhanced}  & 42.6 & 15 & 57.3 & 14 & 14.5 & 5 Rand \\
             $\textup{RSGN}_{unsup}$ & \cite{zhao2021reconstructive} & 42.3 & 16 & 58.0 & 12 & 14 & 5 Rand \\
             UnpairedVSN & \cite{rochan2019video} & 47.5 & 12 & 55.6 & 15 & 13.5 & 5 Rand \\
             PCDL & \cite{zhao2019property} & 42.7 & 14 &  58.4 & 10 & 12 & 5 FCV \\
             ACGAN & \cite{he2019unsupervised} & 46.0 & 13 & 58.5 & 9 & 11 & 5 FCV \\
             $\textup{SUM-}{Ind}_{LU}$ & \cite{yaliniz2021using} &  46.0 & 13 & 58.7 & 8 & 10.5 & $\textup{5 Rand}^\dagger{}$ \\
             ERA & \cite{wu2021era} & 48.8 & 9 & 58.0 & 12 & 10.5 & $\textup{5 Rand}^\dagger{}$ \\
             SUM-GAN-sl & \cite{apostolidis2019stepwise} & 47.8& 11 & 58.4 & 10 & 10.5 & $\textup{5 Rand}^\dagger{}$ \\
             SUM-GAN-AAE & \cite{apostolidis2020unsupervised} & 48.9 & 8 & 58.3 & 11 & 9.5 & $\textup{5 Rand}^\dagger{}$ \\
             $\textup{MCSF}_{late}$ & \cite{kanafani2021unsupervised} & 47.9 & 10 & 59.1 & 6 & 8 &  $\textup{5 Rand}^\dagger{}$ \\
             $\textup{SUM-GDA}_{unsup}$ & \cite{li2021exploring} & 50.0 & 7 & 59.6 & 5 & 6 & 5 FCV \\
             CSNet+GL+RPE & \cite{jung2020global} & 50.2 & 6 & 59.1& 6 & 6 & 5 FCV \\
             CSNet & \cite{jung2019discriminative} & 51.3 & 2 & 58.8 & 7 & 4.5 & 5 FCV \\
             DSR-RL-GRU  & \cite{phaphuangwittayakul2021self} & 50.3& 5 & 60.2 & 4 & 4.5 & $\textup{5 Rand}^\dagger{}$ \\
             AC-SUM-GAN & \cite{apostolidis2020ac} & 50.8 & 4 & 60.6 & 3 & 3.5 & $\textup{5 Rand}^\dagger{}$ \\
             CA-SUM & \cite{apostolidis2022summarizing} & 51.1& 3  & 61.4 & 2 & 2.5 & $\textup{5 Rand}^\dagger{}$ \\ \midrule
             $\textbf{\textup{VVS}}_{\scriptsize{\textit{3840}}}$ \textbf{(ours)} & \multicolumn{1}{c}{-} & \textbf{51.7} & \textbf{1} & \textbf{61.5} & \textbf{1} & \textbf{1} & $\textup{5 Rand}^\dagger{}$  \\
            
            \cmidrule[\heavyrulewidth]{1-8}
            \morecmidrules
            \cmidrule[\heavyrulewidth]{1-8} 
        \end{tabular}
            \caption{\textbf{Benchmark on SumMe and TVSum.} This presents comparisons with unsupervised video summarization methods on SumMe and TVSum. The results for comparison are derived from the previous paper~\cite{apostolidis2022summarizing}. \textit{Rank.} refers to the ranking within these methods, and $\dagger{}$ indicates that its results are evaluated using the same five splits of data.\vspace{-0mm}}
        \label{tab:summarization}
        \end{center}
    \end{table*}

    \subsection{Applicability to Video Summarization} \label{applicability}
        This section presents the further applicability of the proposed suppression scheme, which can be applied not only for video retrieval but also for video summarization. The core objectives of video retrieval and summarization differ significantly; the former is based on vectorization for distinctive representation between unrelated videos, while the latter is based on score regression for human interest. Indeed, to date, there has been no comparison between retrieval methods and summarization methods. However, since our method calculates importance per frame for distinctive vectorization via suppression, we also compared it with summarization methods to demonstrate the scalability of this importance. As shown in~\Cref{tab:summarization}, our method shows state-of-the-art performance compared to other summarization methods. Here, we utilized the same data splits used by existing methods in SumMe~\cite{gygli2014creating} and TVSum~\cite{song2015tvsum} to obtain our results. Furthermore, the comparison is conducted to unsupervised methods since our approach was learned without any annotations, just with a triplet, where a positive is a self-augmented video from an anchor and a negative is a different video. Consequently, the results indicate that the proposed scheme closely aligns with human interest, implying that suppression via explicit signals is one of the solutions towards achieving an optimal summarization.

    \begin{table}[!t] 
        \footnotesize
        \centering
        \begin{tabular}{c|ccc|ccc}
            \cmidrule[\heavyrulewidth]{1-7}
            \morecmidrules
            \cmidrule[\heavyrulewidth]{1-7}
            & \multicolumn{1}{c|}{\textit{Elim.}}  & \multicolumn{2}{c|}{\textit{Gen.}} & \multicolumn{3}{c}{FIVR-5K} \\ \cmidrule(){2-7} 
           
                &\multicolumn{1}{c|}{DDM} & TSM    & TGM    & DSVR  & CSVR  & ISVR  \\ \midrule 
            (a) &       &        &        & 0.605 &	0.615 &	0.580 \\ \midrule
            (b) &\cmark &        &        & 0.626 &	0.637 &	0.598 \\ \midrule
            (c) &       & \cmark &        & 0.611 &	0.620 &	0.585 \\
            (d) &       &        & \cmark & 0.621 &	0.632 &	0.599 \\
            (e) &       & \cmark & \cmark & 0.624 &	0.635 &	0.602 \\ \midrule 
            (f) &\cmark & \cmark &        & 0.632 &	0.641 &	0.603 \\ 
            (g) &\cmark &        & \cmark & 0.635 &	0.646 &	0.619 \\ \midrule
            (h) &\cmark & \cmark & \cmark & \textbf{0.636} & \textbf{0.648} & \textbf{0.620} \\
            \cmidrule[\heavyrulewidth]{1-7}
            \morecmidrules
            \cmidrule[\heavyrulewidth]{1-7}
        \end{tabular}
        \vspace{-1.5mm}
        \caption{\textbf{Module-wise Ablations for $\textup{VVS}_{\small{\textit{500}}}$\textbf{.} }\textit{Elim.} refers to the easy distractor elimination stage, and \textit{Gen.} refers to the suppression weight generation stage. (a) represents a baseline of the same dimension that weighs all frames equally without any of the proposed modules, (b)-(g) represents module-wise ablations, and (h) represents $\textup{VVS}_{\small{\textit{500}}}$.}
        \label{tab:module_ablation500}
    \end{table}
    \subsection{Further Module-wise Ablations} \label{further_abl}
        This section covers further ablation studies for each module in the proposed framework:~$\textup{VVS}_{\small{\textit{500}}}$, $\textup{VVS}_{\small{\textit{512}}}$, and $\textup{VVS}_{\small{\textit{1024}}}$. They exhibit similar tendencies to the module-wise ablation studies of $\textup{VVS}_{\small{\textit{3840}}}$ presented in the main script, as shown in~\Cref{tab:module_ablation500}, \Cref{tab:module_ablation512}, and \Cref{tab:module_ablation1024}. Specifically, within these tables, each module (b)-(d) presents improvements over the baseline (a). In addition, when modules are paired with one another (e)-(g) and when all modules are combined (h), further improvements can be observed. These results suggest that even if the entire framework operates with reduced dimensions, the proposed modules are still effective.

    \subsection{Additional Qualitative Results}\label{qual_exp}
\begin{figure*}[t]
        \centering
        \includegraphics[width=0.99\textwidth]{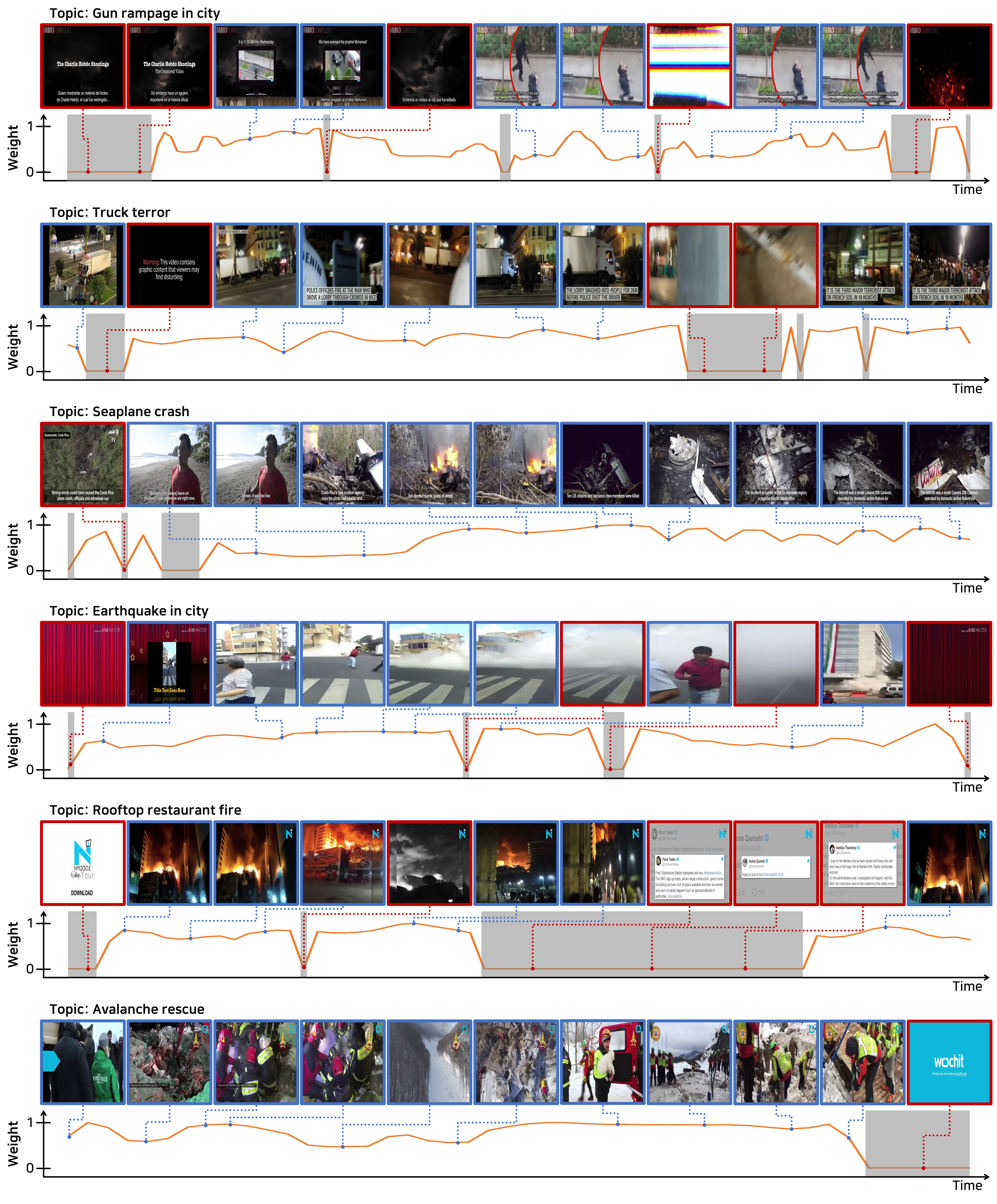} 
        \caption{\textbf{Additional Qualitative Results on FIVR-5K.} The orange line refers to the weights from TSM and TGM; the lower the value, the more suppressed the frame. The gray region corresponds to easy distractors eliminated by DDM, and frames that belong to this area are denoted by a red border.} \label{fig:add_qual}
    \end{figure*}

\begin{figure*}
        \centering
        \subfloat[Training]{{ \includegraphics[width=0.51\textwidth ]{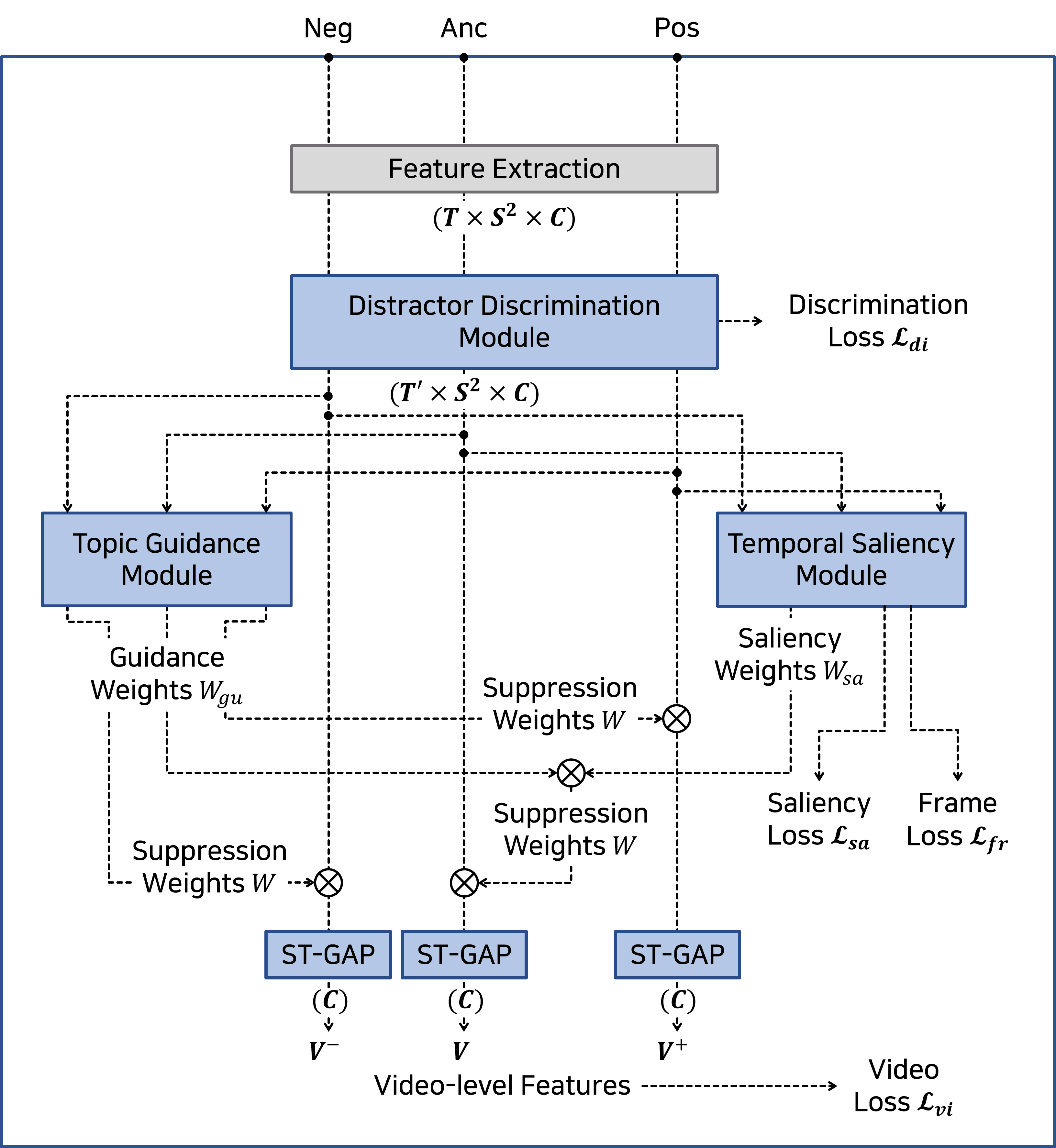} }}%
        \hspace{0.1\textwidth}
        \subfloat[Inference]{{ \includegraphics[width=0.3039\textwidth ]{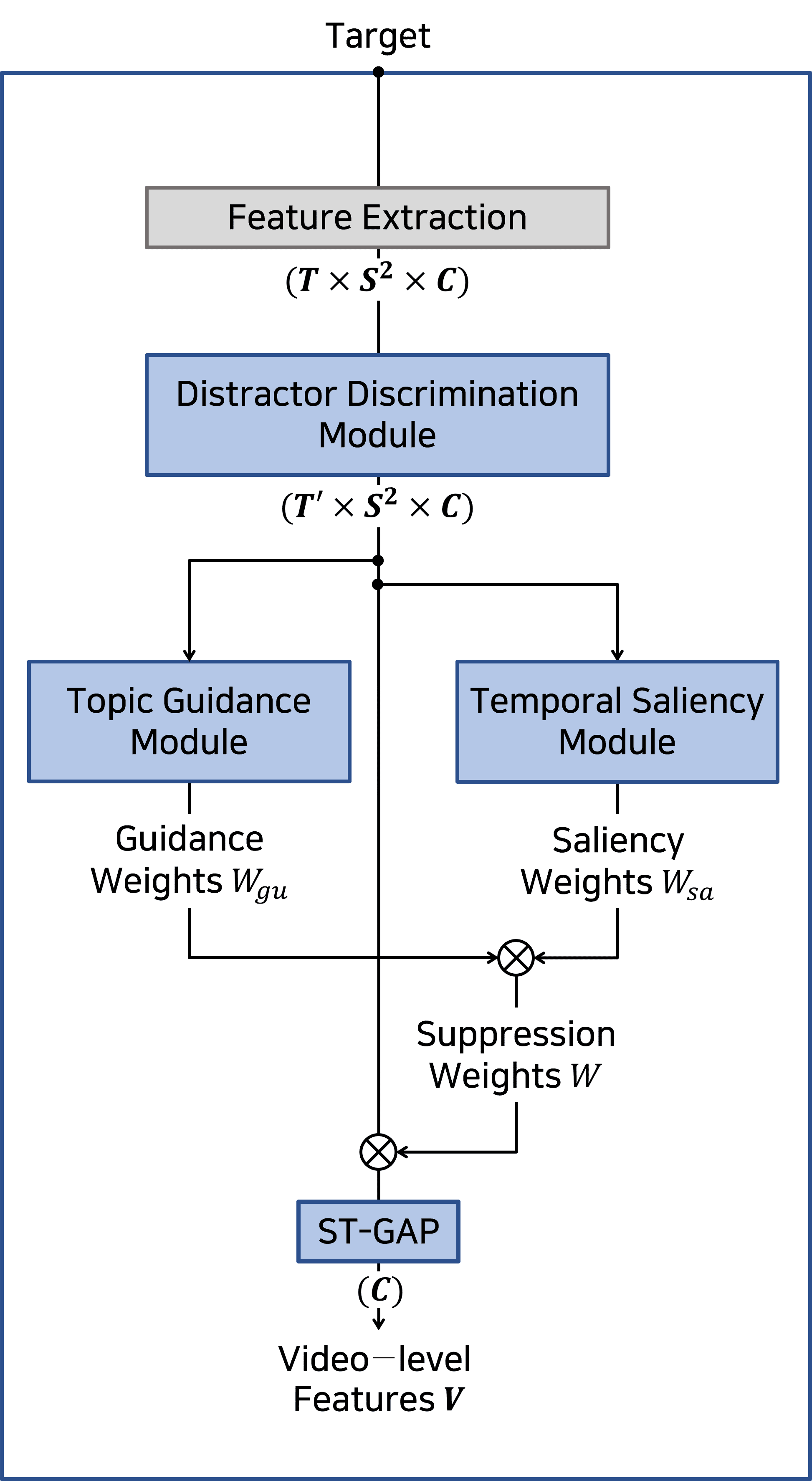} }}%
        \caption{\textbf{Detailed Pipeline of VVS.} $V$, $V^{+}$, and $V^{-}$ refer to the video-level features of anchor, positive, and negative belonging to a triplet, respectively. $\otimes$ refers to the Hadamard product.} \label{fig:detail_pip_vvs}
    \end{figure*}

\begin{figure*}
        \centering
        \subfloat[Training]{{ \includegraphics[width=0.3838\textwidth ]{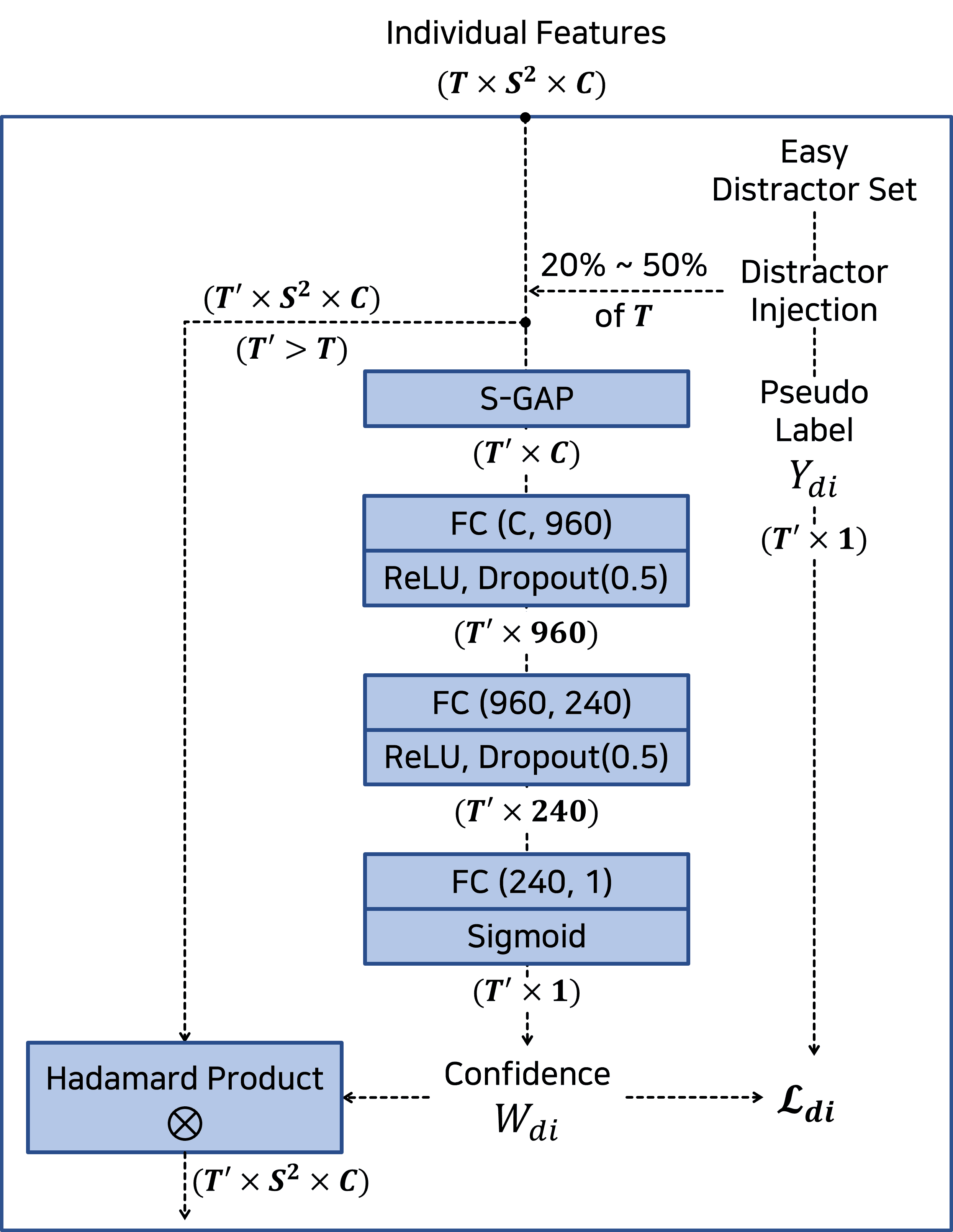} }}%
        \hspace{0.1\textwidth}
        \subfloat[Inference]{{ \includegraphics[width=0.2562\textwidth ]{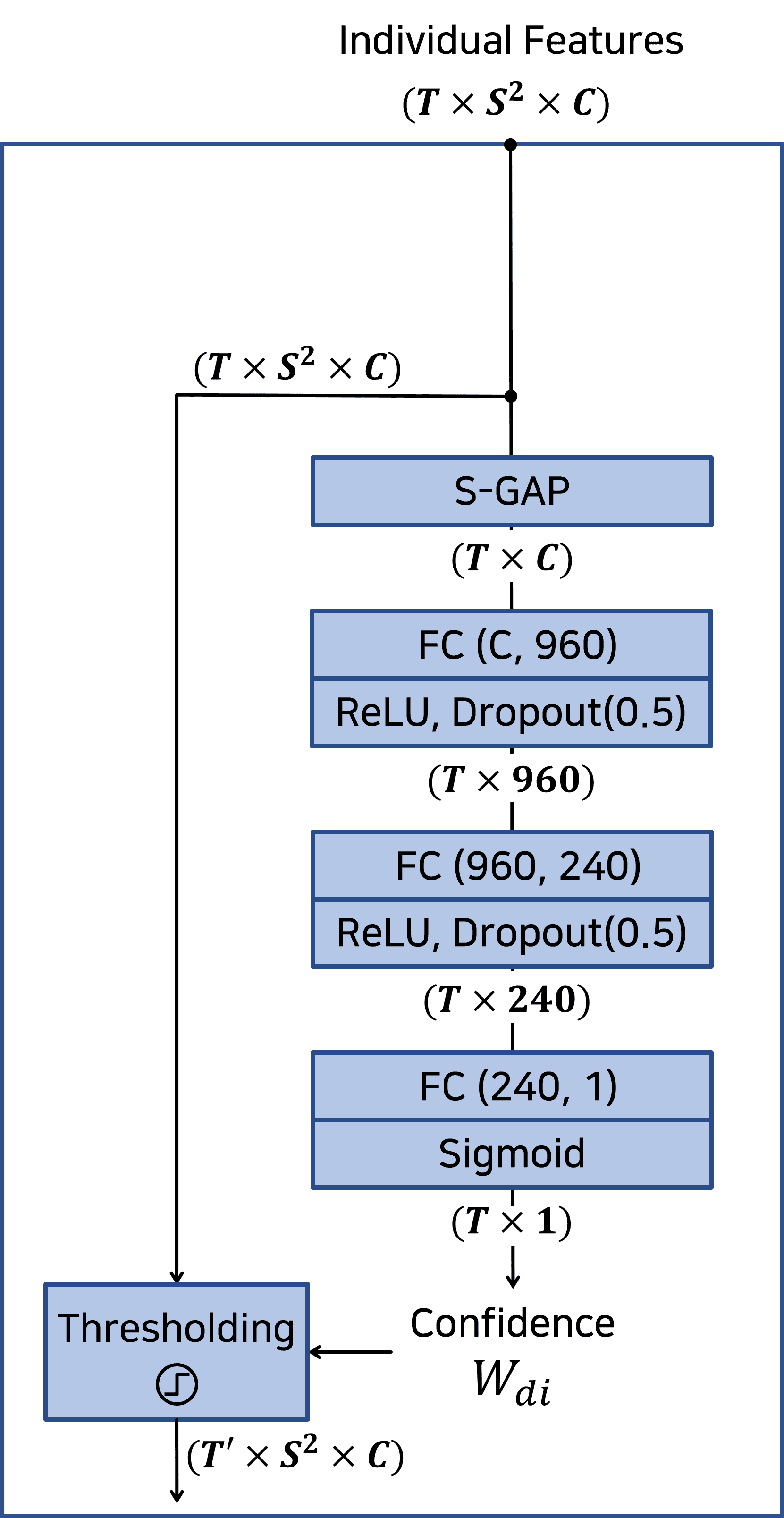} }}%
        \caption{\textbf{Detailed Pipeline of DDM.} The number in parentheses in FC layer blocks indicates the input and output dimensions. The number in parentheses in dropout layer blocks indicates the probability of an element being zeroed.} \label{fig:detail_pip_ddm}
    \end{figure*}

\begin{figure*}
        \subfloat[Training]{{ \includegraphics[width=0.658\textwidth ]{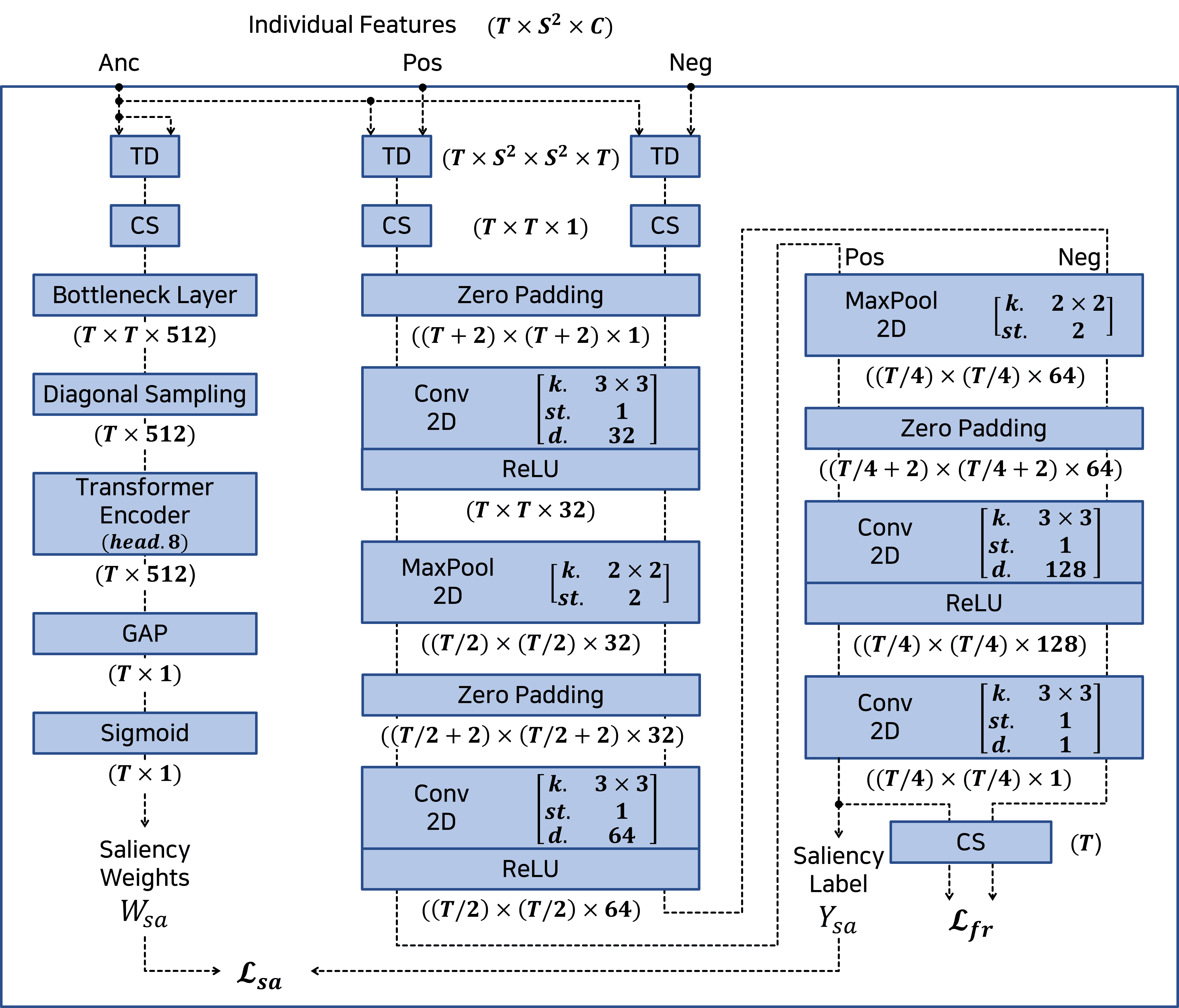} }}%
        \hspace{0.1\textwidth}
        \subfloat[Inference]{{ \includegraphics[width=0.2033\textwidth ]{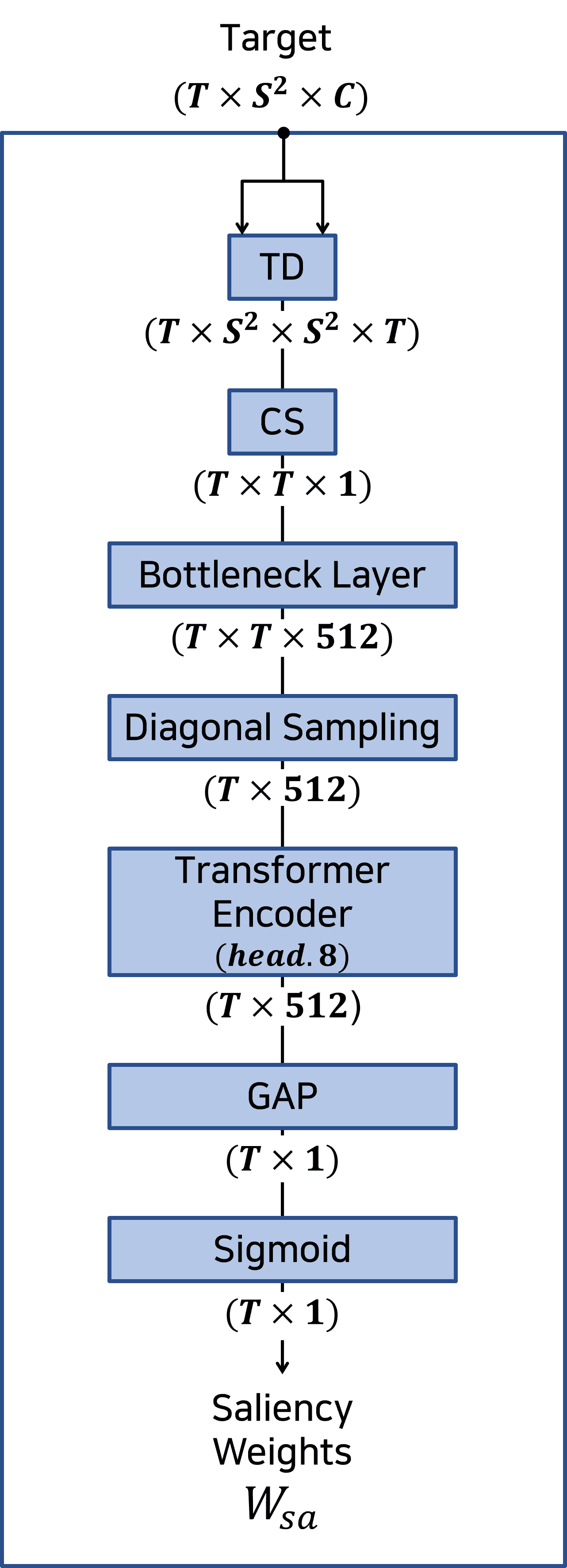} }}%
        \caption{\textbf{Detailed Pipeline of TSM.} The parentheses in each layer block indicate the parameters for that block. \textit{k.} denotes kernel size, \textit{st.} stride, \textit{d.} dimension, and \textit{head.} the number of heads in the multi-head attention within the transformer encoder.} \label{fig:detail_pip_tsm}
    \end{figure*}

\begin{figure*}
        \centering
        \includegraphics[width=0.675\textwidth]{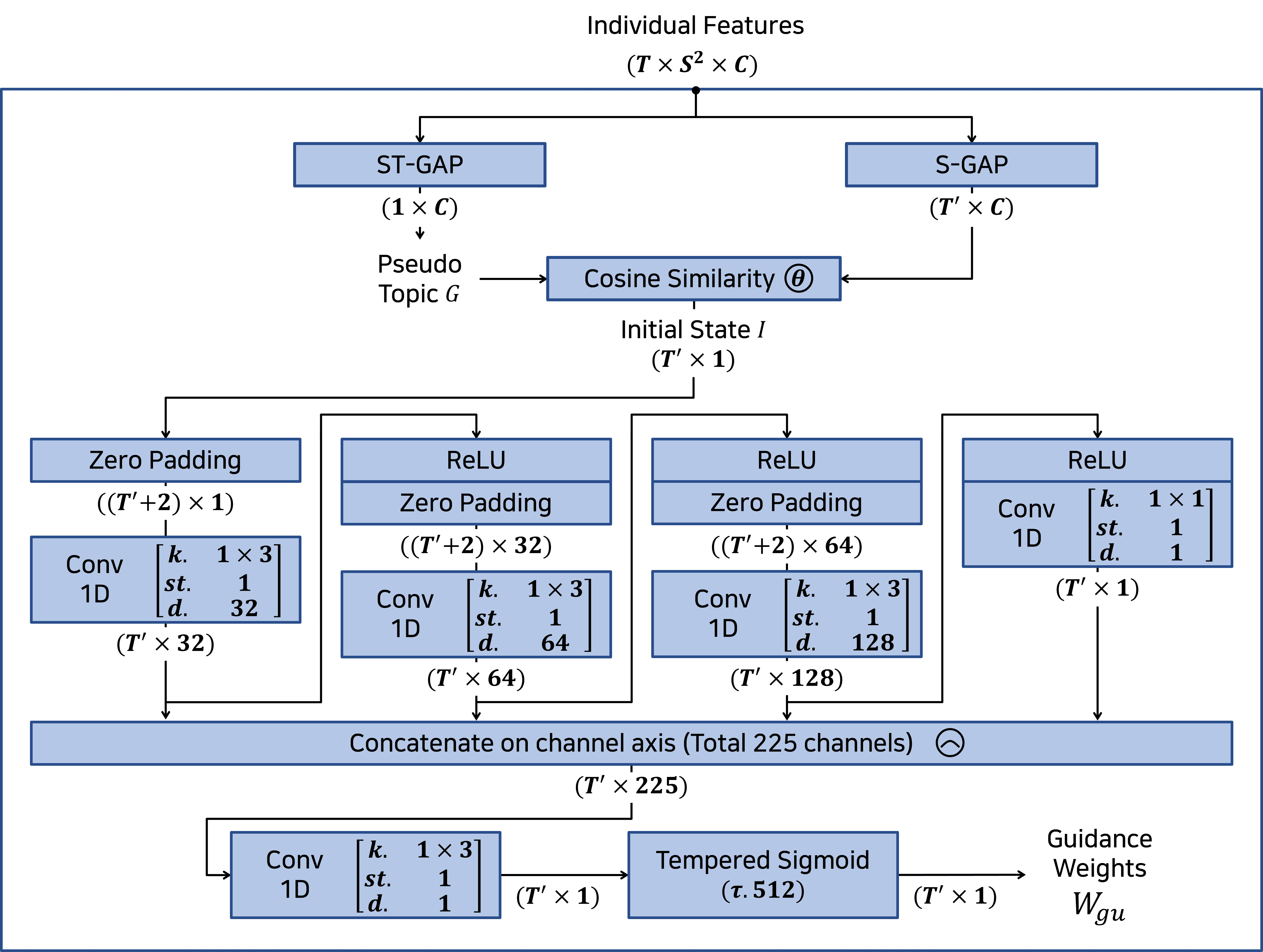} 
        \caption{\textbf{Detailed Pipeline of TGM.} The same pipeline is used for training and inference. The parentheses in each layer block indicate the parameters for that block. \textit{k.} denotes kernel size, \textit{st.} stride, \textit{d.} dimension, and \boldmath{$\tau.$} the temperature of the tempered sigmoid.} \label{fig:detail_pip_tgm}
    \end{figure*}

        This section describes qualitative results on the FIVR-5K in addition to those presented in the main script, as depicted in~\Cref{fig:add_qual} (on page 12). These results reveal that the proposed approach effectively suppresses irrelevant frames in untrimmed videos containing a wide variety of content, as intended in the original design.

        First, when observing the gray areas erased as easy distractors by DDM, it can be seen that they mainly indicate frames with a lack of visual information. Specifically, frames containing simply text or logos are eliminated, such as the first and eighth frames of the video with the topic ``Rooftop restaurant fire". In addition, due to rapid movement during recording, such as in the eighth and ninth frames of the video with the topic ``Truck terror", the complete blur frames are also removed. Furthermore, frames containing limited information, such as the seventh and ninth frames of the video with the topic ``Earthquake in city", are excluded. Because these cases provide insufficient knowledge for recognizing topics, they should be ignored when describing distinct video-level features.

    \begin{table}[!t] 
        \footnotesize
        \centering
        \begin{tabular}{c|ccc|ccc}
            \cmidrule[\heavyrulewidth]{1-7}
            \morecmidrules
            \cmidrule[\heavyrulewidth]{1-7}
            & \multicolumn{1}{c|}{\textit{Elim.}}  & \multicolumn{2}{c|}{\textit{Gen.}} & \multicolumn{3}{c}{FIVR-5K} \\ \cmidrule(){2-7} 
           
                &\multicolumn{1}{c|}{DDM} & TSM    & TGM    & DSVR  & CSVR  & ISVR  \\ \midrule 
            (a) &       &        &        & 0.606 &	0.616 &	0.580 \\ \midrule
            (b) &\cmark &        &        & 0.626 &	0.637 &	0.598 \\ \midrule
            (c) &       & \cmark &        & 0.612 &	0.621 &	0.585 \\
            (d) &       &        & \cmark & 0.623 &	0.633 &	0.600 \\
            (e) &       & \cmark & \cmark & 0.625 &	0.635 &	0.602 \\ \midrule 
            (f) &\cmark & \cmark &        & 0.632 &	0.643 &	0.603 \\ 
            (g) &\cmark &        & \cmark & 0.638 &	0.647 &	0.619 \\ \midrule
            (h) &\cmark & \cmark & \cmark & \textbf{0.643} & \textbf{0.654} & \textbf{0.625} \\
            \cmidrule[\heavyrulewidth]{1-7}
            \morecmidrules
            \cmidrule[\heavyrulewidth]{1-7}
        \end{tabular}
        \vspace{-1.5mm}
        \caption{\textbf{Module-wise Ablations for $\textup{VVS}_{\small{\textit{512}}}$\textbf{.} }\textit{Elim.} refers to the easy distractor elimination stage, and \textit{Gen.} refers to the suppression weight generation stage. (a) represents a baseline of the same dimension that weighs all frames equally without any of the proposed modules, (b)-(g) represents module-wise ablations, and (h) represents $\textup{VVS}_{\small{\textit{512}}}$. \vspace{-1mm}}
        \label{tab:module_ablation512}
    \end{table}
    
    \begin{table}[t!] 
        \footnotesize
        \centering
        \begin{tabular}{c|ccc|ccc}
            \cmidrule[\heavyrulewidth]{1-7}
            \morecmidrules
            \cmidrule[\heavyrulewidth]{1-7}
            & \multicolumn{1}{c|}{\textit{Elim.}}  & \multicolumn{2}{c|}{\textit{Gen.}} & \multicolumn{3}{c}{FIVR-5K} \\ \cmidrule(){2-7} 
           
                &\multicolumn{1}{c|}{DDM} & TSM    & TGM    & DSVR  & CSVR  & ISVR  \\ \midrule 
            (a) &       &        &        & 0.642 &	0.650 &	0.609 \\ \midrule
            (b) &\cmark &        &        & 0.662 &	0.672 &	0.628 \\ \midrule
            (c) &       & \cmark &        & 0.645 &	0.653 &	0.616 \\
            (d) &       &        & \cmark & 0.654 &	0.665 &	0.631 \\
            (e) &       & \cmark & \cmark & 0.656 &	0.666 &	0.635 \\ \midrule 
            (f) &\cmark & \cmark &        & 0.668 &	0.676 &	0.636 \\ 
            (g) &\cmark &        & \cmark & 0.672 &	0.682 &	0.653 \\ \midrule
            (h) &\cmark & \cmark & \cmark & \textbf{0.678} & \textbf{0.688} & \textbf{0.652} \\
            \cmidrule[\heavyrulewidth]{1-7}
            \morecmidrules
            \cmidrule[\heavyrulewidth]{1-7}
        \end{tabular}
        \vspace{-1.5mm}
        \caption{\textbf{Module-wise Ablations for $\textbf{\textup{VVS}}_{\small{\textit{1024}}}$\textbf{.} }\textit{Elim.} refers to the easy distractor elimination stage, and \textit{Gen.} refers to the suppression weight generation stage. (a) represents a baseline of the same dimension that weighs all frames equally without any of the proposed modules, (b)-(g) represents module-wise ablations, and (h) represents $\textup{VVS}_{\small{\textit{1024}}}$. \vspace{0mm}}
        \label{tab:module_ablation1024}
    \end{table}
    
        Moreover, when observing the suppression weights generated by TSM and TGM, the scales of the weights fluctuate based on the degree to which each frame is related to the topic. Specifically, high weight values are assigned because they give a meaningful clue even if the region involved in the topic is in picture-in-picture form, such as the third and fourth frames of the video with the topic ``Gun rampage in city". Additionally, even when a specific person is interviewed about the topic, such as in the second and third frames of the video with the topic ``Seaplane crash", it is assigned low weight values because the event pertaining to the topic is not visually displayed and may be seen in other topic videos. Furthermore, frames that are associated with the topic but do not explicitly depict a rescue activity, such as the fifth frame of the video with the topic ``Avalanche rescue", are given a relatively low weight, whereas frames in which the activity occurs are given high weights. From the above instances, it is clear that the suppression weights determine the amount to which a frame is excluded depending on how directly it connects to the event indicated by the topic for each frame.

\end{document}